\documentclass[review,times]{elsarticle}
\usepackage{amsmath}
\usepackage{graphicx}
\usepackage{subfig} 
\usepackage{tabu}  
\usepackage{multirow}
\usepackage{dsfont}
\usepackage{stfloats}
\usepackage{booktabs}
\usepackage{algorithm}
\usepackage{algorithmic}
\usepackage{amssymb}
\usepackage{mathtools}
\usepackage{caption}
\usepackage{mhchem}
\usepackage{tensor}
\usepackage[colorlinks,linkcolor=blue,anchorcolor=blue,citecolor=green]{hyperref}
\usepackage[a4paper,left=2.5cm,right=2.5cm,top=2.5cm,bottom=2.5cm]{geometry}

\journal{ISPRS Journal of Photogrammetry and Remote Sensing}

\biboptions{authoryear}

\begin{document}

\begin{frontmatter}

\title{Towards Deep and Efficient: A Deep Siamese Self-Attention Fully Efficient Convolutional Network for Change Detection in VHR Images}

\author[mymainaddress]{Hongruixuan Chen}

\author[mymainaddress]{Chen Wu\corref{mycorrespondingauthor}}
\cortext[mycorrespondingauthor]{Corresponding author}
\ead{chen.wu@whu.edu.cn}

\author[mysecondaryaddress]{Bo Du}

\address[mymainaddress]{State Key Laboratory of Information Engineering in Surveying, Mapping, and Remote Sensing, Wuhan University, Wuhan, China.}
\address[mysecondaryaddress]{School of Computer Science, Wuhan University, Wuhan, China.}

\begin{abstract}
  Change detection in multi-temporal very-high-resolution (VHR) images plays a significant role in facilitating the understanding of the relationships and interactions between human activities and the natural environment. Recently, fully convolutional networks (FCNs) have attracted widespread attention in the change detection field. In pursuit of better change detection performance, it has become a tendency to design deeper and more complicated FCNs, which inevitably brings about huge numbers of parameters and an unbearable computational burden. With the goal of designing a quite deep architecture to obtain more precise change detection results while simultaneously decreasing parameter numbers to improve efficiency, in this work, we present a very deep and efficient change detection network, entitled EffCDNet. In EffCDNet, to reduce the numerous parameters associated with deep architecture, an efficient convolution consisting of depth-wise convolution and group convolution with a channel shuffle mechanism is introduced to replace standard convolutional layers. In terms of the specific network architecture, EffCDNet does not use mainstream UNet-like architecture, but rather adopts another type of architecture with a very deep encoder and a lightweight decoder. In the very deep encoder, two very deep siamese streams stacked by efficient convolution first extract two highly representative and informative feature maps from input image-pairs. By performing a subtraction operation on two deep feature maps, a difference feature map containing rich change information is produced. Subsequently, an efficient atrous spatial pyramid pooling (EASPP) module is designed to capture multi-scale change information. In the lightweight decoder, a recurrent criss-cross self-attention (RCCA) module is applied to efficiently utilize non-local similar feature representations to enhance discriminability for each pixel, thus effectively separating the changed and unchanged regions. Moreover, to tackle the optimization problem in confused pixels, two novel loss functions based on information entropy are presented. On two challenging open change detection datasets, our approach outperforms other state-of-the-art FCN-based methods in terms of both visual interpretation and accuracy assessment, with only benchmark-level parameter numbers and quite low computational overhead, thereby demonstrating its effectiveness, superiority, and potential.
\end{abstract}

\begin{keyword}
    Change detection, very-high-resolution image, deep siamese fully convolutional network, efficient convolution, self-attention mechanism, information entropy
\end{keyword}

\end{frontmatter}

\section{Introduction}
\par Remote sensing techniques can offer large-scale, long-term, and periodic observations over the ground surface \citep{Bovolo2015, Bergen2019}.Using multi-temporal remote sensing images to detect earth surface changes (i.e., change detection) has become a hot topic in the remote sensing field \citep{Singh1989, Zhu2017b, Liu2019}. Due to the rapid development of Earth observation technology, numerous optical sensors (e.g., IKONOS, QuickBird, GaoFen, and WorldView) have been developed that are capable of providing a large number of multi-temporal images with high spatial resolution, which has expanded the potential applications of change detection. How to utilize these massive VHR images for precise change detection is significant for fields including land-cover and land-use change analysis, urban planning and development, precision agriculture, cadastral survey, and damage assessment \citep{Xian2009,Luo2018,Shi2020b,Brunner2010,Vetrivel2018,Wu2018}. A large amount of literature has been published over the past decades with a focus on change detection \citep{Hussain2013, Sharma2007, Nielsen1997,Wu2014,Bovolo2008, Liu2015b, Hoberg2015, Lei2014}. However, the obvious geometric structures and complex texture information in VHR images pose significant challenges for these traditional methods; this is because these methods only explore “shallow” features and many of them are hand-crafted, which is both unrepresentative and shows poor robustness \citep{wu2019unsupervised}. Under these circumstances, deep learning was introduced to facilitate the extraction of high-level, hierarchical, and representative features for change detection. 

\par Recent years have witnessed the success of deep learning in a wide range of fields, including computer vision, natural language processing, and remote sensing image interpretation \citep{Lecun2015,Zhang2016b,Zhu2017a,Liu2020c}. A series of change detection methods have accordingly been developed based on deep models \citep{Gong2017b, Du2019a, Chen2019, Chen2019a}. Gong \emph{et al.} \citep{Gong2017b} used a superpixel segmentation-based method to generate reliable samples, and further designed a DBN for difference representation learning in VHR images. Following recent developments in multi-scale feature learning \citep{Szegedy2015a}, Chen \emph{et al.} \citep{Chen2019} presented a deep siamese multi-scale convolutional network and a corresponding pre-detection algorithm for change detection in VHR images. Regarding the change detection task as a sequential prediction problem, Lyu \emph{et al.} \citep{Lyu2016} utilized an improved long short-term memory (LSTM) to learn the land-cover change rules. By combining CNN and RNN, Mou \emph{et al.} \citep{Mou2019} developed an end-to-end convolutional recurrent neural network to learn spatial-spectral-temporal features for binary and multi-class change detection. All the above well-established methods are patch-wise models, in which a corresponding neighbor region is first generated for each pixel in multi-temporal images, after which they are passed into deep models for feature extraction and change detection. Since each pixel is represented by a fixed-size patch, these models have a limited and fixed receptive field, which restricts their change detection performance. Moreover, the patch generation step introduces redundant computational cost; for large-scale multi-temporal images, the space and time costs brought in by this step are enormous \citep{CayeDaudt2018, Xu2020}. 

\par To overcome the drawbacks of the patch-wise method, Daudt \emph{et al.} \citep{CayeDaudt2018} introduced the fully convolutional network (FCN) for the change detection task and presented two siamese FCN architectures. Due to replacing fully connected layers with 1$\times$1 convolutional layers, the FCN architecture can take multi-temporal images of arbitrary size and directly produce the complete change maps without the patch generation step, thereby achieving better performance and higher efficiency \citep{Shelhamer2017}. Following the work of Daudt \emph{et al.}, many change detection studies have been carried out based on FCN architecture. Lei \emph{et al.} \citep{Lei2019} proposed an FCN with a symmetric U-shape for landslide inventory mapping, in which a pyramid pooling module \citep{Zhao2017} is utilized to capture the multi-scale change information. For the same multi-scale feature extraction purpose, Chen \emph{et al.} \citep{Chen2019, chen2020change} developed a deep siamese multi-scale FCN based on a multi-scale feature convolution unit and utilized fully connected conditional random field (FC-CRF) to balance the local and global change information. Inspired by the UNet++ architecture proposed for medical images, Peng \emph{et al.} \citep{Peng2019a} presented an improved UNet++ (IUNet++), which is capable of learning multi-level features in VHR images and achieves decent change detection performance in an open-source dataset. In addition, by comprehensively utilizing pre-training, deep supervision, and attention mechanism, Zhang \emph{et al.} \citep{Zhang2020} presented a novel image fusing network (IFN) that can produce accurate change detection results. 

\par After considering the above-mentioned existing works, it can be concluded that in order to obtain better change detection performance, building deeper and more complicated networks has become a primary tendency. However, this tendency has resulted in a sharp increase in both parameter numbers and computational cost. From FC-EF \citep{CayeDaudt2018}, the first FCN model presented for change detection, to IFN \citep{Zhang2020}, one of the state-of-the-art methods, the parameter numbers have increased by 20.8 times, and the computational cost has increased by 18 times. Although the hardware technology has developed rapidly, the computational cost of these state-of-the-art methods remains a serious obstacle for real-time processing of large-scale and massive multi-temporal data from various Earth observations on portable devices. Therefore, how to balance the conflict between performance improvement and computational burden brought by deep architecture is worth considering. In terms of the specific network architecture, most previous FCN-based methods have simply adopted UNet-like architecture. The most obvious characteristic of these networks is that the encoder and decoder networks are symmetric. For change detection, considering the diversity of changed objects and scenes, the encoder is required to extract representative and informative deep features from multi-temporal images to the great extent possible. Subsequently, the decoder simply needs to separate the changed and unchanged pixels to generate change maps. The importance of the encoder and decoder in change detection tasks is not equal; hence, a symmetric structure may lead to insufficient parameters for the encoder, resulting in poor performance, and excessive parameters for the decoder, generating extra computational resource requirements without equivalent performance improvement. In addition, there exist some “confused pixels” that are difficult to optimize for cross-entropy loss; for example, a few non-building changes in the building change detection task and some pseudo-changes caused by seasonal variations. Due to the lack of sufficient optimization in the training stage, these pixels often have a similar probability of change and non-change, i.e., high uncertainty, causing false and missing alarms in the final result. 

\par These issues motivate us to propose a very deep and efficient end-to-end change detection network (EffCDNet) for change detection. The network depth of EffCDNet reaches 116 layers while maintaining benchmark-network-level parameter numbers and low computational overhead. Our main contributions can be summarized as follows: 

\begin{enumerate} 
  \item This paper presents the first attempt at designing a very deep but lightweight change detection network. Compared with other state-of-the-art methods, the proposed network achieves better performance on two open large-scale datasets with only benchmark-level numbers of parameters and computational overhead.
  \item To balance the conflict between performance boost and the computational burden brought about by deep architecture, an efficient convolution consisting of depth-wise convolution, group point-wise convolution, and channel shuffle is used to replace the standard convolution. The application of this efficient convolution allows us to deepen our network without incurring a high degree of computational cost.
  \item The self-attention mechanism is introduced for change detection tasks to separate changed and unchanged pixels. To further decrease the computational cost and make the network more efficient, a recurrent criss-cross attention module is applied rather than the standard self-attention module.
  \item In order to further optimize confused pixels in change detection tasks, two novel information entropy-based loss functions are presented; this enables better optimization of confused pixels in the training stage, thus enhancing the network’s performance.
\end{enumerate}

\par The remainder of this paper is organized as follows. Section \ref{sec:2} elaborates on the proposed network in detail. In Section \ref{sec:3}, the experimental results and related discussions conducted on two open datasets are presented. Finally, the conclusion of our work is drawn in Section \ref{sec:4}. 

\section{Methodology}\label{sec:2}
\par In this section, we first elaborate on the efficient convolution that factorizes standard convolution into the combination of depth-wise convolution, group point-wise convolution, and channel shuffle. Based on the efficient convolution, we propose the very deep and efficient network EffCDNet and describe the key modules of our network in detail, including residual channel shuffle (RCS) unit, efficient ASPP module, RCCA module, and information entropy-based loss function. 

\subsection{Efficient Convolution}
\begin{figure*}[t]
  \centering
  \subfloat[]{
    \includegraphics[scale=0.7]{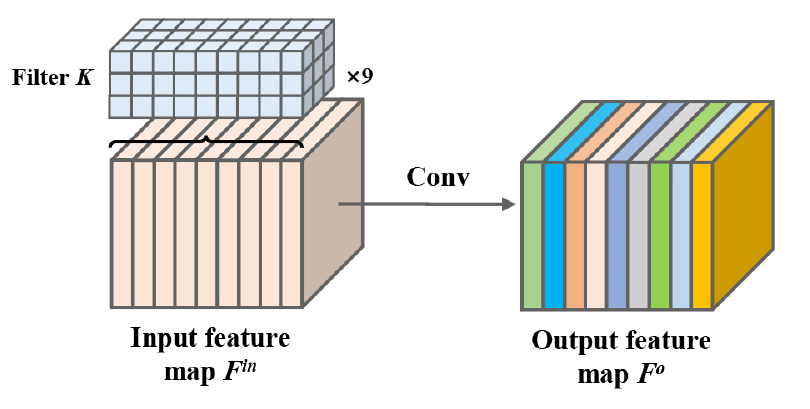}
  \label{fig:CSCD}
  }
  \subfloat[]{
    \includegraphics[scale=0.7]{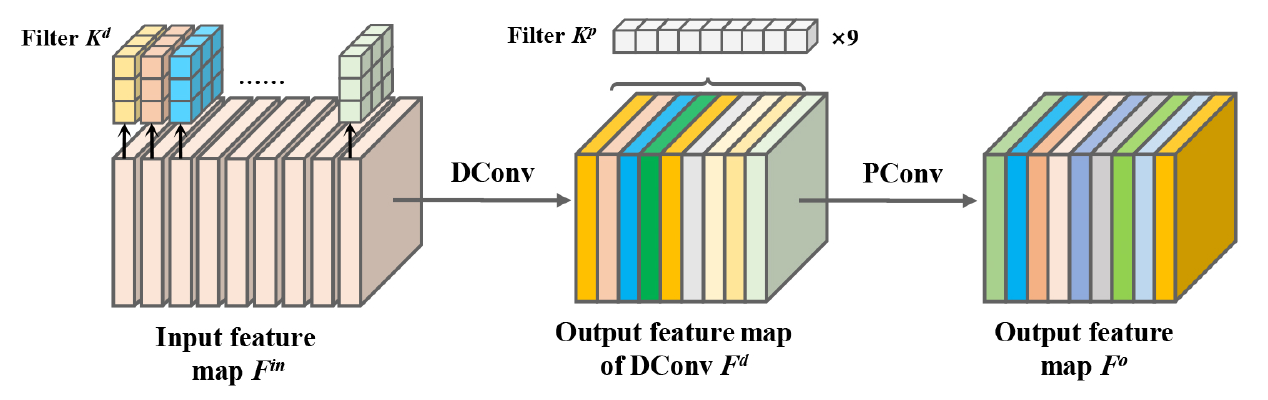}
  \label{fig:moti}
  }
  \caption{The illustrations of (a) standard convolution, and (b) depth-wise separate convolution. }
  \label{fig:1}
\end{figure*}
\par As change detection networks become deeper and more complicated, the standard convolutional layer introduces an increasingly large number of parameters, thereby incurring high computational and storage costs. Considering an input feature map $F^{in}\in R^{H\times W\times C_{in}}$ and a produced output feature map $F^{o}\in R^{H\times W\times C_{o}}$, a standard convolutional layer (Fig. 1-(a)) parameterized by filters $K\in R^{S_{k}\times S_{k}\times C_{in}\times C_{o}}$ can be expressed as follows:

\begin{equation}
  \begin{aligned}
    F^{o}(h, w, c) &=\left(F^{i n} * K\right)(h, w, c) \\
    &=\sum_{i=1}^{S_{k}} \sum_{j=1}^{S_{k}} \sum_{n=1}^{C_{i n}} K(i, j, n, c) F^{i n}(h+i, w+j, n).
    \end{aligned}
\label{eq:1}
\end{equation}
If $C_{in}=C_{o}=512$, $S_{k}=3$, a standard convolutional layer would have 2.36 million trainable parameters. In practice, many of these deep layers would be stacked, thereby causing huge parameter numbers. It is therefore necessary to explore efficient variants of convolution for change detection tasks. 

\par Depth-wise separate convolution \citep{ioffe2015batch, howard2017mobilenets} is a widely used structure to reduce the network parameters, as shown in Fig. 1-(b), by splitting the standard convolution into a depth-wise convolution and a 1$\times$1 convolution, called point-wise convolution. In depth-wise convolution parameterized with filters $K^{d}\in R^{S_{k}\times S_{k}\times 1\times C_{in}}$, each kernel extracts features on only one channel: 
\begin{equation}
\begin{aligned}
  F^{d}(h, w, c) &=\left(F^{i n} * K^{d}\right)(h, w, c) \\
  &=\sum_{i=1}^{S_{k}} \sum_{j=1}^{S_{k}} K^{d}(i, j, 1, c) F^{i n}(h+i, w+j, c).
  \end{aligned}
  \label{eq:2}
\end{equation}
\par However, each channel of feature maps produced by depth-wise convolution is isolated from every other channel. To solve the problem of no information interaction occurring between channels, following depth-wise convolution, a point-wise convolution with filters $K^{p}\in R^{1\times 1\times C_{in}\times C_{o}}$ is used to fuse the information in each channel:
\begin{equation}
\begin{aligned}
  F^{o}(h, w, c) &=\left(F^{d} * K^{p}\right)(h, w, c) \\
  &=\sum_{n=1}^{C_{i n}} K^{p}(1,1, n, c) F^{d}(h, w, n)
  \end{aligned}
  \label{eq:3}
\end{equation}
The above process, which factorizes standard convolution into two steps, can reduce the parameter numbers to $1 / C_{o}+1 / S_{k}^{2}$ of the standard convolution.

\begin{figure*}[t]
    \centering
    \subfloat[]{
      \includegraphics[scale=0.67]{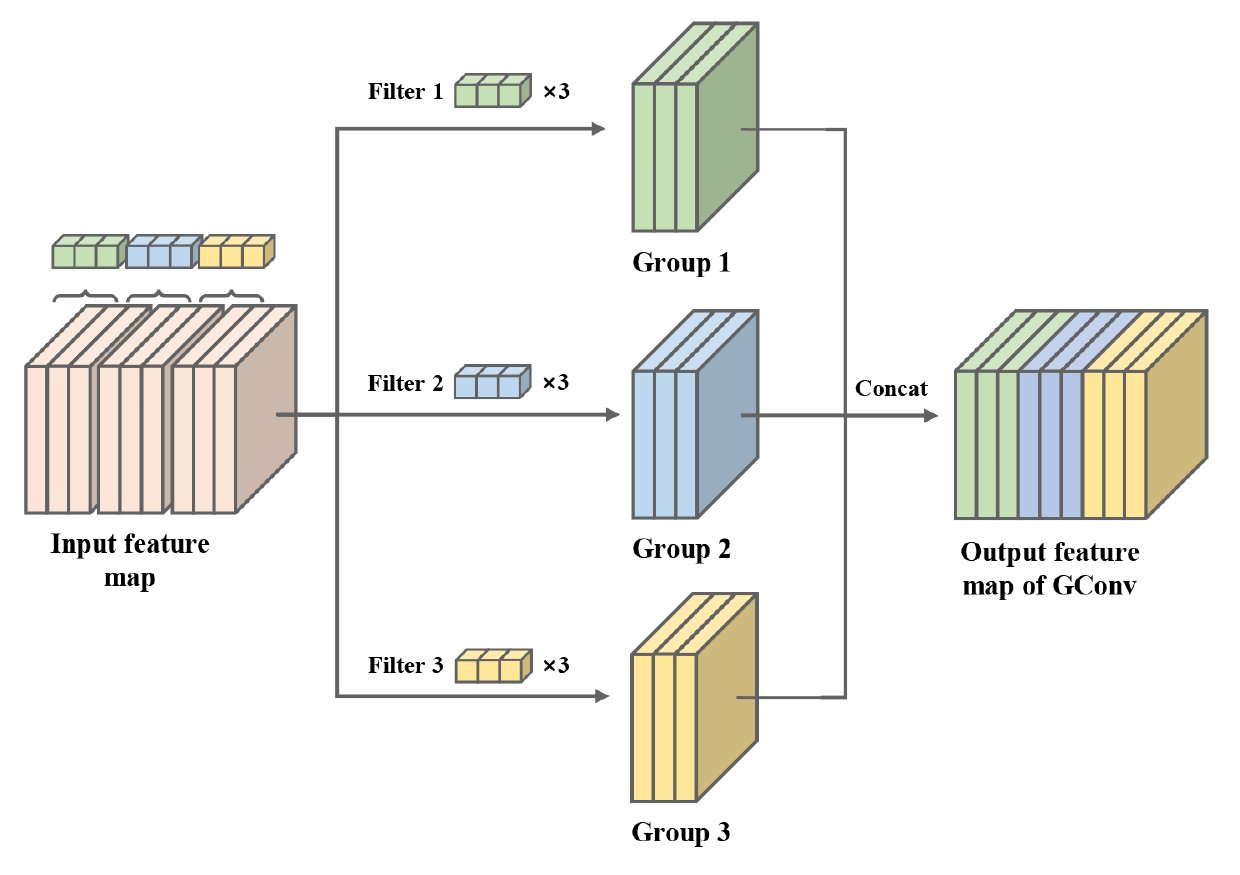}
    \label{fig:CSCD}}
    \hfil
    \begin{minipage}[b]{0.45\textwidth}
      \centering
      \subfloat[]{
      \includegraphics[scale=0.47]{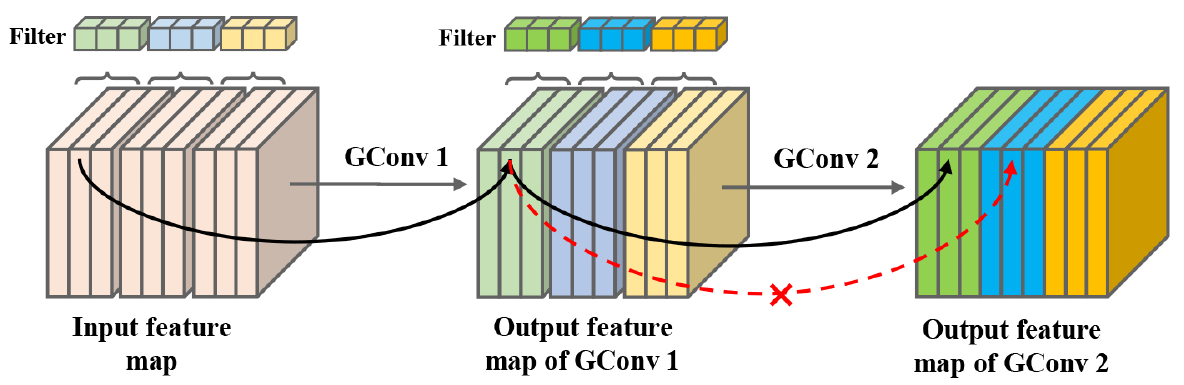}
      }
      \hfil
      \subfloat[]{
      \includegraphics[scale=0.47]{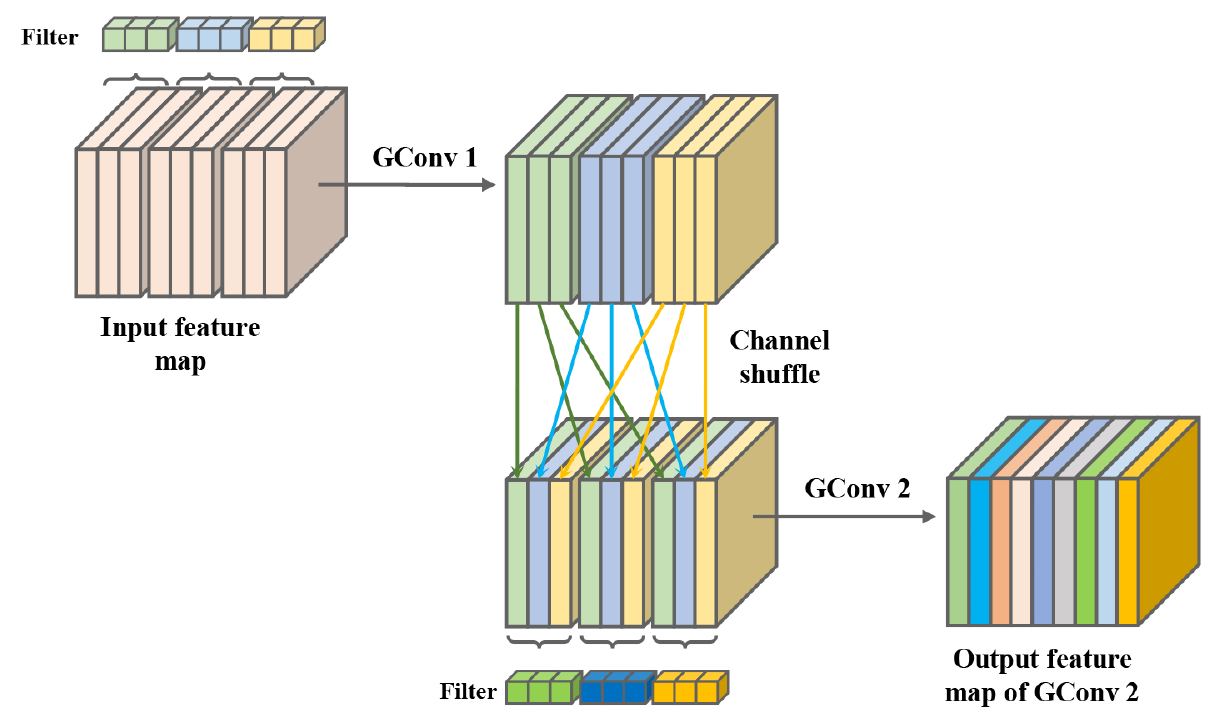}
      }
    \end{minipage}
    \caption{The illustrations of (a) group point-wise convolution, (b) information isolation problem in group convolution, and (c) group point-wise convolution with channel shuffle.}
    \label{fig:2}
\end{figure*}


\par After the standard convolution has been replaced by depth-wise separate convolution, point-wise convolution occupies the majority of the parameters. More specifically, in a depth-wise separate convolution with $S_{K}=3$, $C_{in}=512$, and $C_{o}=512$, point-wise convolution contributes 98.3$\%$ of the number of parameters. Therefore, to further reduce parameters and construct more efficient models, group point-wise convolution \citep{Xie2017} is introduced. In group point-wise convolution, the input feature map and convolution filters are divided into $G$ groups: 
\begin{equation}
\left\{\begin{array}{l}
  K^{p}=\left[K_{1}^{p}, K_{2}^{p}, \cdots, K_{G}^{p}\right], K_{g}^{p} \in R^{1 \times 1 \times\left(C_{in} / G\right) \times\left(C_{o} / G\right)} \\
  F^{i n}=\left[F_{1}^{i n}, F_{2}^{i n}, \cdots, F_{G}^{i n}\right], F_{g}^{i n} \in R^{H \times W \times\left(C_{in} / G\right)}
  \end{array}\right. .
  \label{eq:4}
\end{equation}
Then, the filters $K^{p}_{g}$ of group $g$ only operate on the corresponding $g$-th input feature maps $F^{in}_{g}$: 
\begin{equation}
\begin{aligned}
  F_{g}^{o}(h, w, c) &=\left(F_{g}^{i n} * K_{g}^{p}\right)(h, w, c) \\
  &=\sum_{n=1}^{C_{i n} / G} K_{g}^{p}(1,1, n, c) F_{g}^{i n}(h, w, n).
\end{aligned}
\label{eq:5}
\end{equation}
Subsequently, by concatenating the output feature map $F^{o}_{g}$ of each group in channel, we can obtain the final output feature map $F^{o}=\left[F_{1}^{o}, F_{2}^{o}, \cdots, F_{G}^{o}\right] \in R^{H \times W \times C_{o}}$. By means of the grouping operation, the parameter numbers of point-wise convolution are reduced to $1/G$. Fig. 2-(a) presents the diagram of group point-wise convolution.

\par Nevertheless, considering that multiple group point-wise convolutions are stacked together, an information isolation problem exists between groups. As illustrated in Fig. 2-(b), it is obvious that the output produced by a certain group relates only to the input within the corresponding group, and that there is no information interaction between groups, which might seriously damage the network performance. To address this issue, we utilize a channel shuffle mechanism \citep{Zhang2018q} for cross-group information exchange. More specifically, for an output feature map of point-wise group convolution $F^{o} \in R^{H \times W \times C_{o}}$, it is first reshaped in the channel dimension $F_{re}^{o} \in R^{H \times W \times G \times\left(C_{o} / G\right)}$. The last two dimensions of $F_{re}^{o}$ are then transposed and flattened back into the original channel dimension $\tilde{F}^{o} \in R^{H \times W \times C_{o}}$. With the help of this simple channel shuffle mechanism, the information deposited in each group is reassigned, as shown in Fig. 2-(c). 

\begin{figure*}[t]
  \centering
  \includegraphics[scale=0.51]{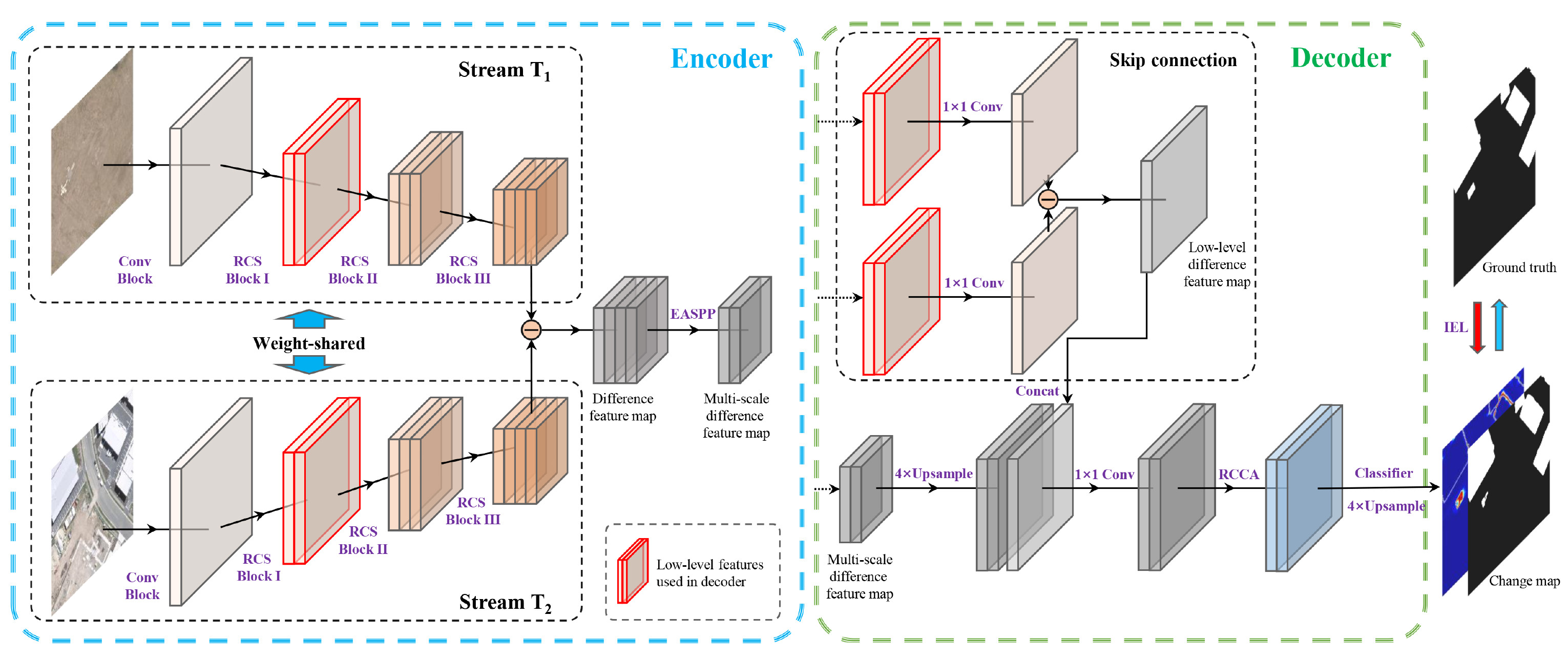}
  \caption{The network architecture of the proposed EffCDNet. The encoder network consists of two very deep siamese streams and one efficient ASPP module. The key components of the decoder network are a skip connection structure, a RCCA module and a classifier. The encoder network receives two multi-temporal VHR images and outputs one multi-scale difference feature map. The decoder network combines the multi-scale difference feature map with a low-level difference feature map, which is generated by the two low-level feature maps output by RCS block I. The feature representation becomes more discriminative by means of the RCCA module. Subsequently, the classifier outputs the change map. Finally, the parameters of the whole network are optimized using the information entropy-based loss function. }
  \label{fig:3}
\end{figure*}

\par Based on the aforementioned depth-wise convolution, point-wise group convolution and channel shuffle mechanism, we can build a powerful and efficient deep network for change detection tasks.

\subsection{Network Architecture}
\par Utilizing the efficient convolution as the basic unit, the overall network architecture of our proposed EffCDNet is presented in Fig. \ref{fig:3}. The encoder of EffCDNet is designed to be very deep, enabling it to extract representative and informative deep features in order to cope with the complex ground situations of VHR images. However, in addition to a large number of parameters, a very deep network often suffers from training problems such as gradient vanishing and the derogation problem \citep{He2016}. Accordingly, a RCS unit that combines residual learning and efficient convolution is utilized to deepen the encoder of our network. The RCS unit is introduced in section 2.2.1. More specifically, the encoder network of EffCDNet, with its two streams $S^{T_{1}}$ and $S^{T_{2}}$, takes a multi-temporal image-pair $I^{T_{1}}$ and $I^{T_{2}}$ as input. To ensure that the deep features extracted from two images are comparable (in the same feature space) and to reduce parameters, the two streams are designed as siamese architectures that have the same structure and are weight-shared. Each stream consists of a standard convolutional block and three RCS blocks. First, the low-level feature representation is abstracted from the input image by means of the standard convolutional block. Three RCS blocks stacked by multiple RCS units then progressively extract more and more abstract and global feature representations. The network depth of each stream reaches 102 layers. The constructed stream is deep enough to enable full extraction of informative and representative deep features from the input image. Despite its depth, the stream suffers from neither training problems nor large parameter numbers owing to the presence of the RCS unit. After yielding two deep feature map $F^{T_{1}}$ and $F^{T_{2}}$ from input images by two siamese streams, a pixel-wise subtraction operation is performed on $F^{T_{1}}$ and $F^{T_{2}}$ for generating deep difference feature map $D=F^{T_{1}}-F^{T_{2}}$, where change information gets highlighted. In the next step, to better handle changed objects with different sizes in a low computational cost way, an EASPP module is applied to $D$ to produce multi-scale difference feature map $D^{ms}$. The specific structure of EASPP module is illustrated in section 2.2.2.

\par For the decoder of EffCDNet, unlike in the existing UNet-like change detection architecture, we argue that the decoder network in change detection tasks does not necessarily have to be as deep and complicated as the encoder: as long as the encoder network can fully extract informative and representative features, and the features of two classes can be adequately separated from each other, a shallow decoder can also recover changed objects very well. Guided by this argument, the decoder of EffCDNet is simple and lightweight. As shown in Fig. \ref{fig:3}, in the decoder network, the multi-scale difference feature map $D^{ms}$ is first upsampled by a factor of 4. Since $D^{ms}$ contains more abstract and global change information, yet less concrete and local change information, some concrete and local information is required to generate changed objects with accurate boundaries. Considering that the low-level feature maps in early layers contain rich local information, the two low-level feature maps generated by the first RCS block are utilized. The two feature maps are reduced dimension in channel axis by two 1$\times$1 convolutional layers, which can avoid the case in which the importance of low-level features outweighs the importance of $D^{ms}$. Through a subtraction operation, a low-level difference feature map $D^{l}$ is generated. $D^{ms}$ and $D^{l}$ are then concatenated and passed through a 1$\times$1 convolutional layer to fuse the low-level and high-level change information. Next, to further separate the changed and unchanged regions, a RCCA module, which employs a self-attention mechanism to use non-local similar feature representation for improving feature discriminability, is utilized to effectively enlarge the feature distance between changed and unchanged pixels. The motivation and description of the RCCA module are presented in section 2.2.3. Although the decoder of EffCDNet is shallow and simple, implementing the procedures outlined above enables the generated features to have both global and local changed information and exhibit high inter-class discriminability. It is therefore easy for the classifier to generate precise change detection results with sharp boundaries. At the end of decoder, we apply a classifier comprising several efficient convolutions and a standard 1$\times$1 convolutional layer, followed by an upsampling by a factor of 4, to produce the prediction result.  

\par Finally, with reference to the prediction result and ground truth, a loss function is applied to optimize the network parameters. However, in change detection tasks, there exist some confused pixels, such as non-building changes in building change detection tasks and pseudo-changes caused by seasonal variation. Cross-entropy loss function, the most widely used loss function in change detection, cannot optimize these pixels very well. To tackle this problem, we instead use Shannon information entropy to measure the uncertainty of each pixel in the prediction map and propose two information entropy-based loss functions. As the confused pixels show high entropy values, they will receive more attention in the training stage. More details about the information entropy-based loss function are provided in section 2.2.4. 

\begin{figure}[t]
  \centering
  \subfloat[]{
    \includegraphics[scale=0.5]{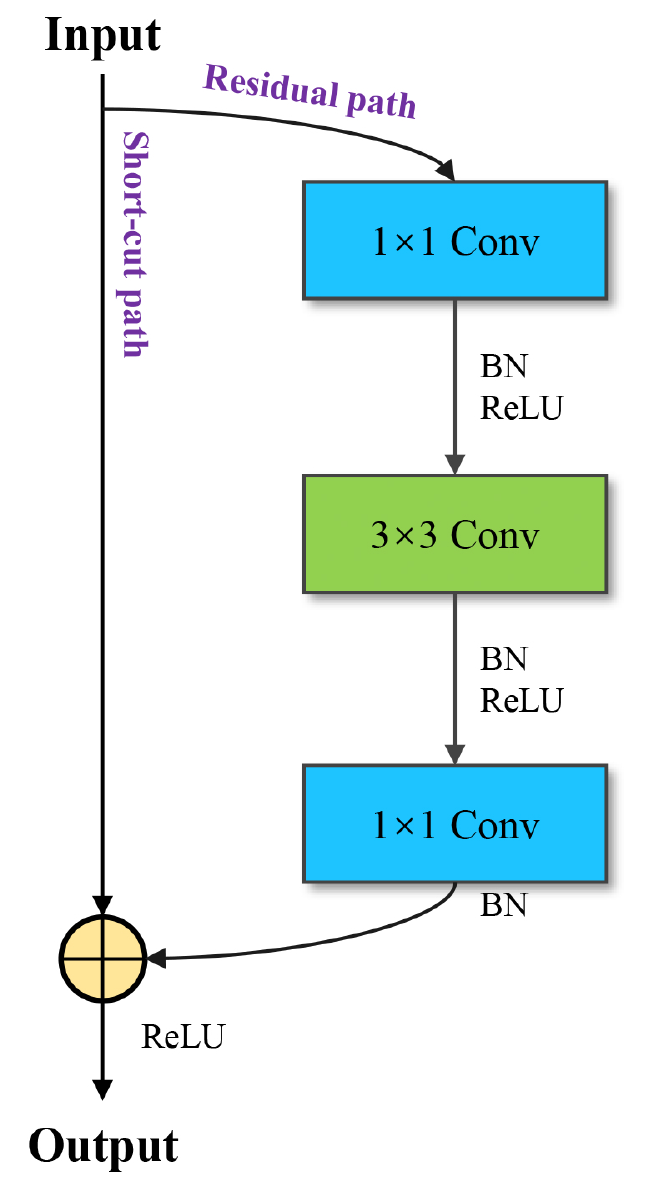}
  \label{fig:CSCD}}
  \hfil
  \subfloat[]{
    \includegraphics[scale=0.5]{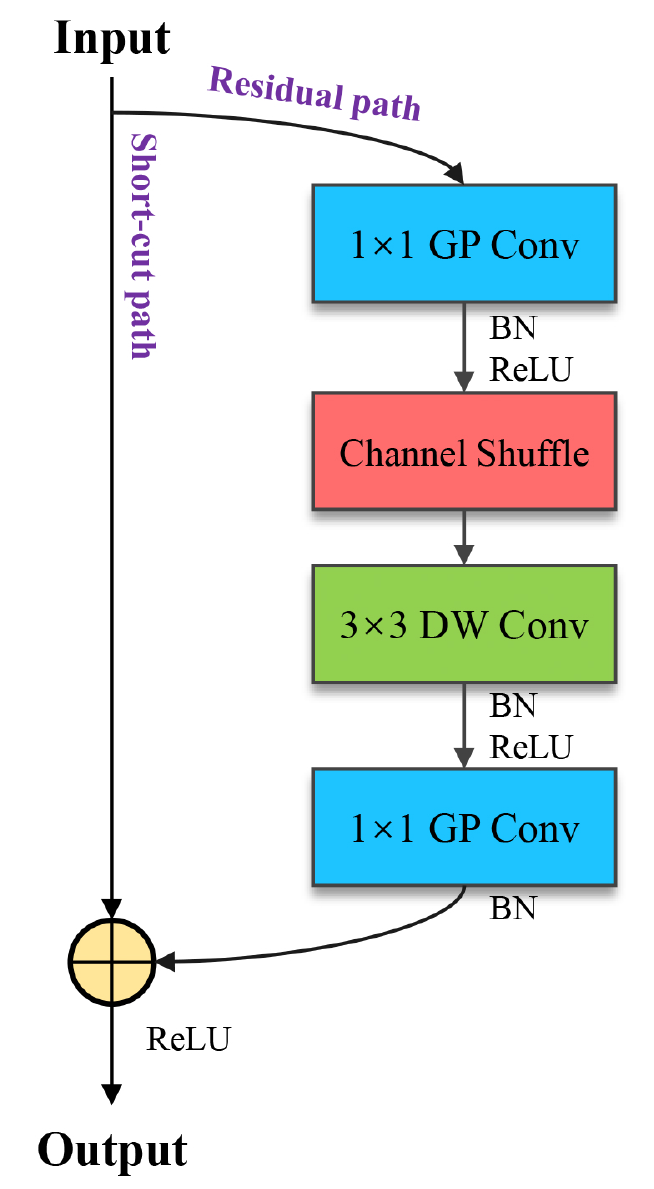}
  \label{fig:CSCD}}
  \hfil
  \subfloat[]{
    \includegraphics[scale=0.5]{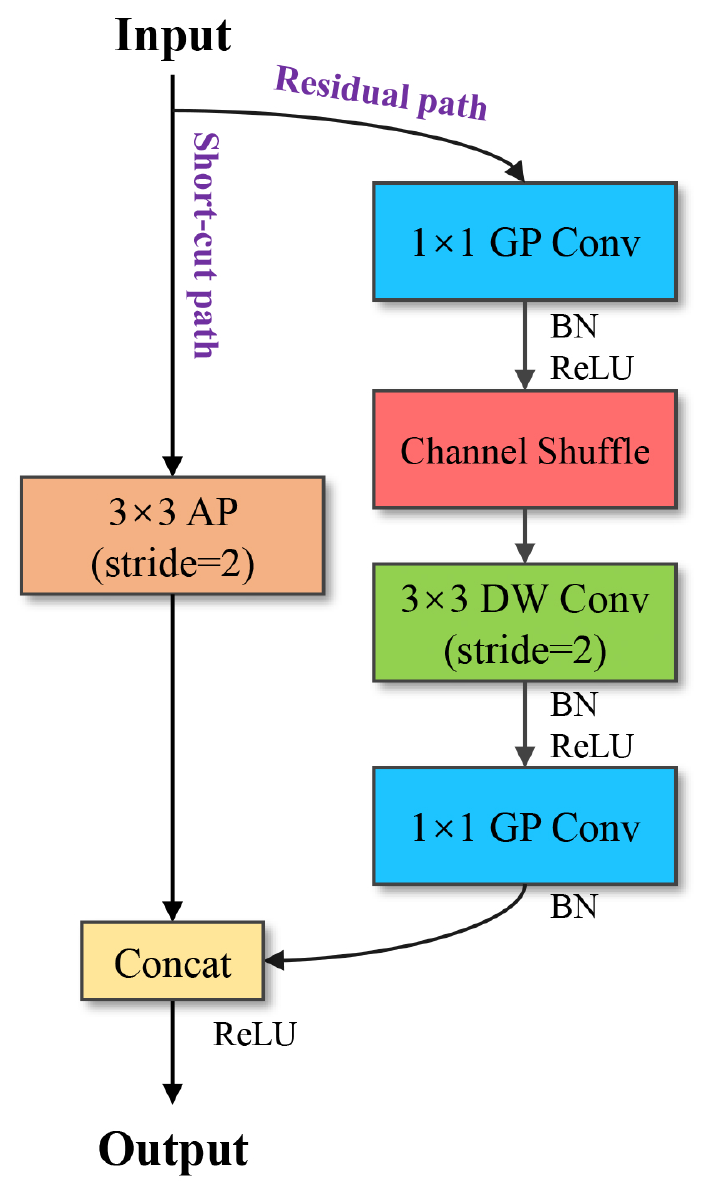}
  \label{fig:moti}}
  \caption{The illustrations of (a) standard residual unit, (b) residual channel shuffle unit, (c) residual channel shuffle unit with stride=2 for down-sampling.}
  \label{fig:4}
\end{figure}

\subsubsection{Residual Channel Shuffle Unit for Very Deep Encoder}
\par In order to fully extract informative and representative features from input VHR images, we aim to design a very deep encoder for change detection. However, constructing a deep encoder network is not as easy as simply stacking more layers. In addition to the large model size, a very deep encoder will face gradient vanishing and degradation problems during the training stage \citep{Ioffe2015, He2016}, which will hamper the network convergence, damage the final performance, and even result in lower accuracy than shallow networks if left unaddressed.

\par Moreover, limited by the training problems, a lot of change detection networks are only about 20 to 30 layers in size, which restricts these networks to achieve better performance. To deal with the training problems, we introduce residual learning into our network \citep{He2016}. A standard residual learning module is illustrated in Fig. 4-(a). However, although the residual learning module is able to solve the training problems of very deep networks, all convolutional layers used in the residual module are standard convolution; hence, it is inevitable that this will introduce numerous parameters as the network depth increases. Therefore, depth-wise convolution, group point-wise convolution, and channel shuffle are adopted to modify the residual module in order to create the residual channel shuffle (RCS) unit. As shown in Fig. 4-(b), in the RCS unit, the two 1$\times$1 convolutional layers are replaced by group point-wise convolution, meaning that their parameters are reduced to $1/G$, (where $G$ is the group number). To avoid the information block problem of group convolution, the first group point-wise convolution is followed by the channel shuffle mechanism to reassign the information in each group. Subsequently, depth-wise convolution is applied to replace the standard 3$\times$3 convolutional layer, thus further reducing parameters. For the down-sampling step, to retain more information, an RCS unit with stride is utilized instead of directly using a pooling operation. Fig. 4-(c) depicts the RCS unit with a stride of 2. In the shortcut path, an average pooling layer with kernel size 3$\times$3 and stride of 2 is applied. In the residual path, a 3$\times$3 depth-wise convolution with a stride of 2 is used to reduce feature size. The down-sampled raw feature map and down-sampled residual feature map are then concatenated to produce the final output. 

\par Benefiting from residual learning and efficient convolution, the RCS unit allows us to construct a very deep encoder for EffCDNet while avoid the obstacles associated with deep architecture. 

\begin{figure}[!t]
  \centering
  \includegraphics[scale=0.56]{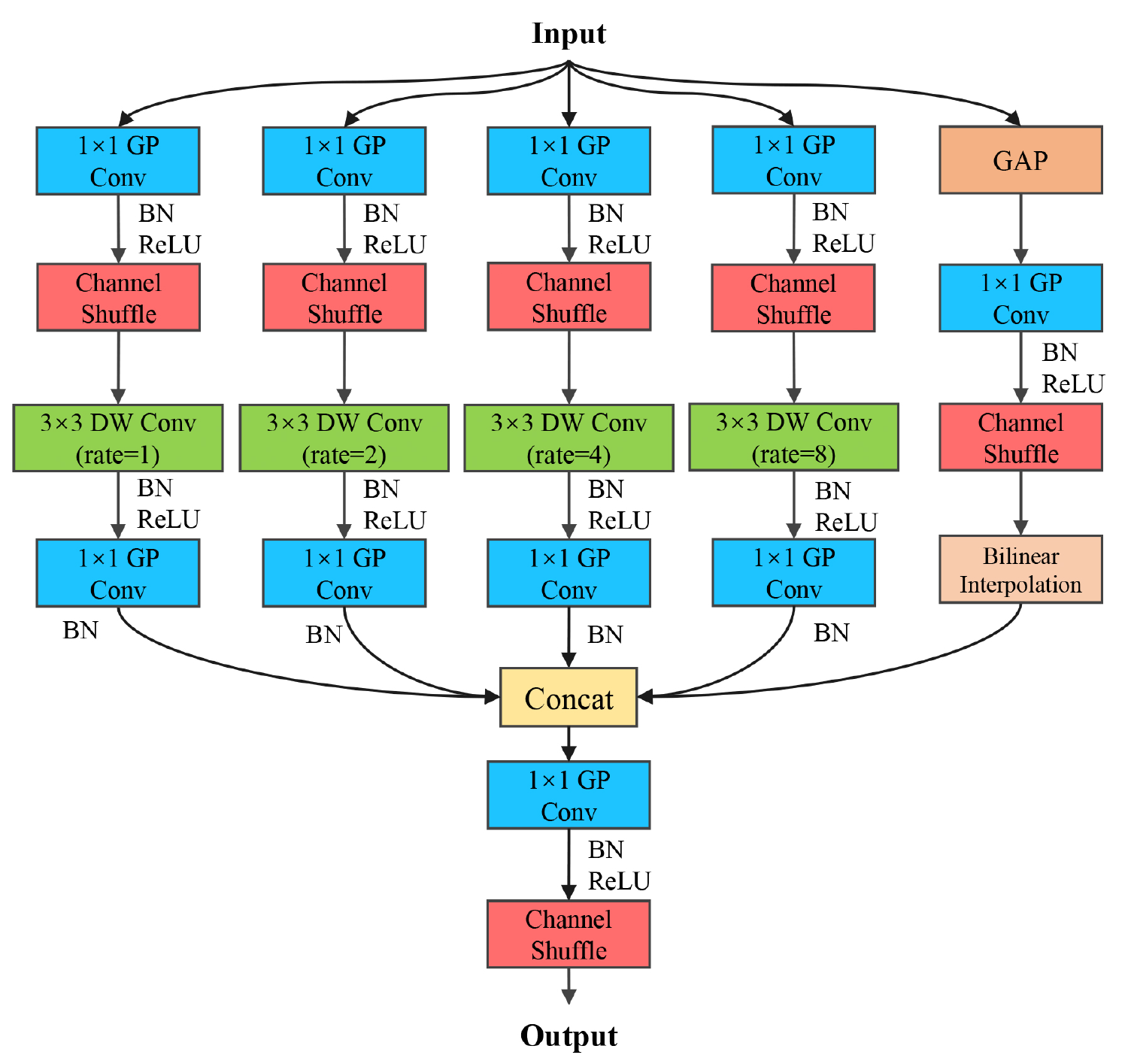}
  \caption{The illustration of EASPP module.}
  \label{fig:5}
\end{figure}

\subsubsection{Efficient ASPP Module for Multi-scale Feature Learning}
\par In the remote sensing images, there exist different ground objects with diverse sizes ranging from tiny neighbors to large regions, which in turn results in various changed objects with multiple scales. To handle multi-scale changed objects, an efficient ASPP (EASPP) module is proposed in EffCDNet. The ASPP module \citep{Chen2018b} is an effective multi-scale feature extraction module, which extracts multi-scale information by means of multiple parallel atrous convolution layers \citep{chen2014semantic} with different dilation rates. Nonetheless, the standard ASPP module performs feature extraction on deep features with high dimensions, resulting in huge parameter numbers and high computational overhead. To address this issue, we propose an efficient ASPP module for learning multi-scale difference representation. Fig. \ref{fig:5} illustrates the proposed EASPP module, which comprises four atrous convolution branches with different dilated rates and one pooling branch. To reduce the number of parameters, in the EASPP module, the standard atrous convolution is decomposed into a combination of depth-wise atrous convolution, group convolution, and channel shuffle. The dilation rates of the four depth-wise atrous convolutions are set to 1, 2, 4, and 8 to effectively extract features at different scales. In the pooling branch, a global pooling layer followed by a 1$\times$1 group convolution with channel shuffle can capture the global information. Finally, the features containing change information on different scales are concatenated and passed through another 1$\times$1 group convolution with channel shuffle to generate the final multi-scale difference features. Notably, in the proposed EffCDNet, the EASPP module has only 0.44 million parameters; in comparison, a standard ASPP module would have 7.70 million parameters. The use of the EASPP module notably reduces the parameters and accordingly allows our network to extract multi-scale difference features with high efficiency.

\begin{figure*}[t]
  \centering
  \begin{minipage}[b]{0.5\textwidth}
    \centering
    \subfloat[]{
    \includegraphics[scale=0.28]{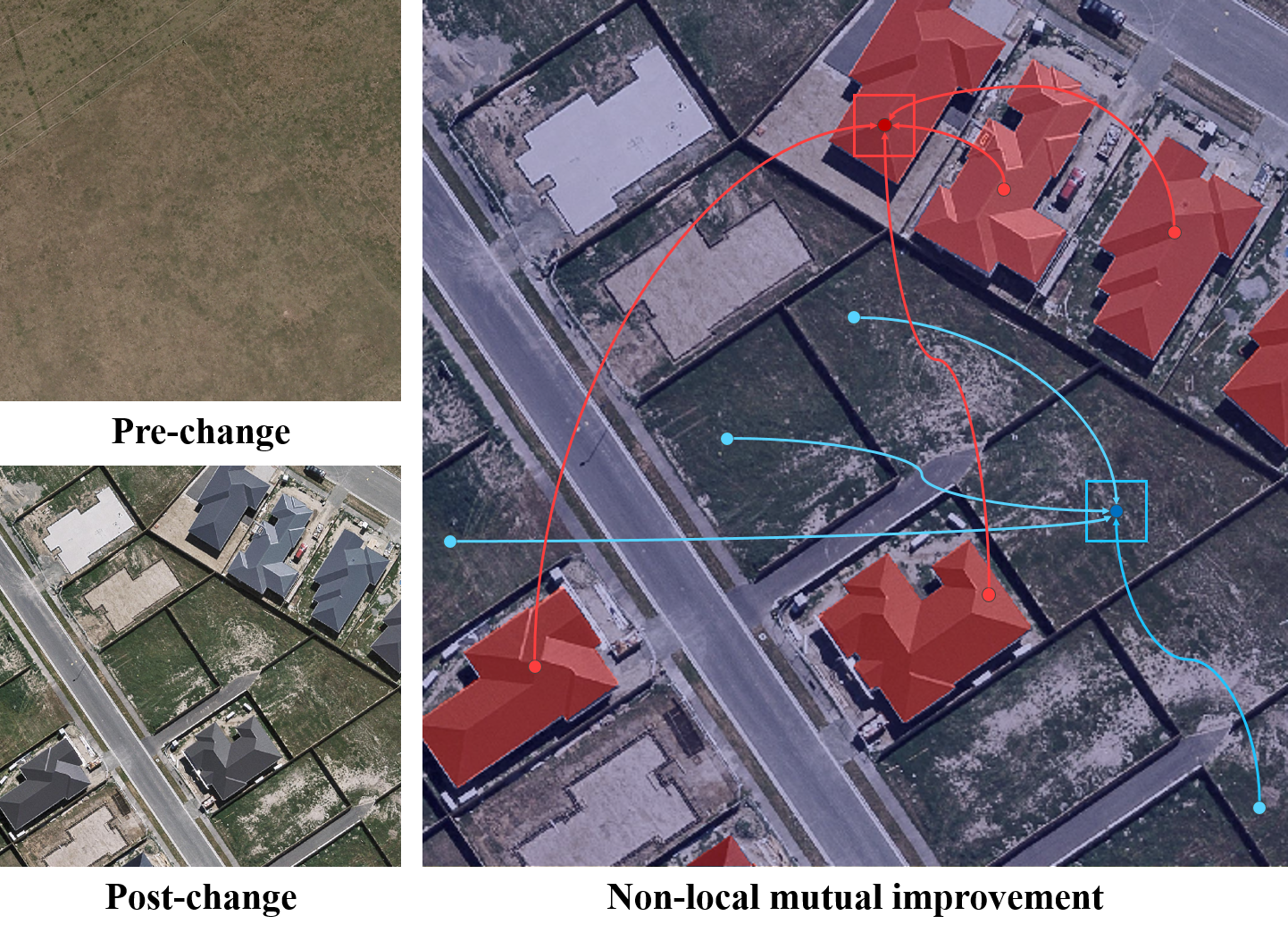}
    }
    \hfil
    \subfloat[]{
    \includegraphics[scale=0.28]{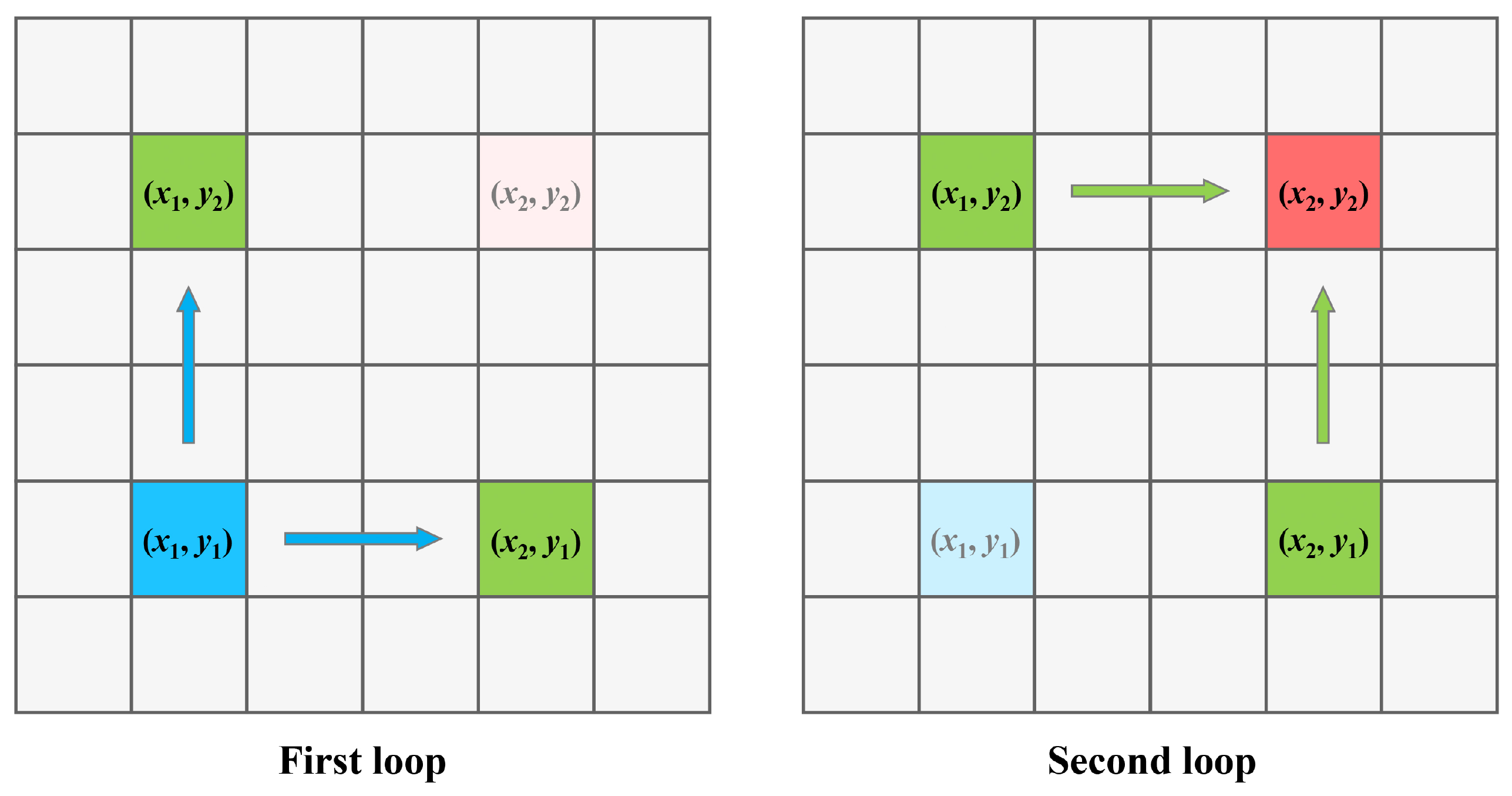}
    }
  \hfil
  \end{minipage}
  \subfloat[]{
    \includegraphics[scale=0.36]{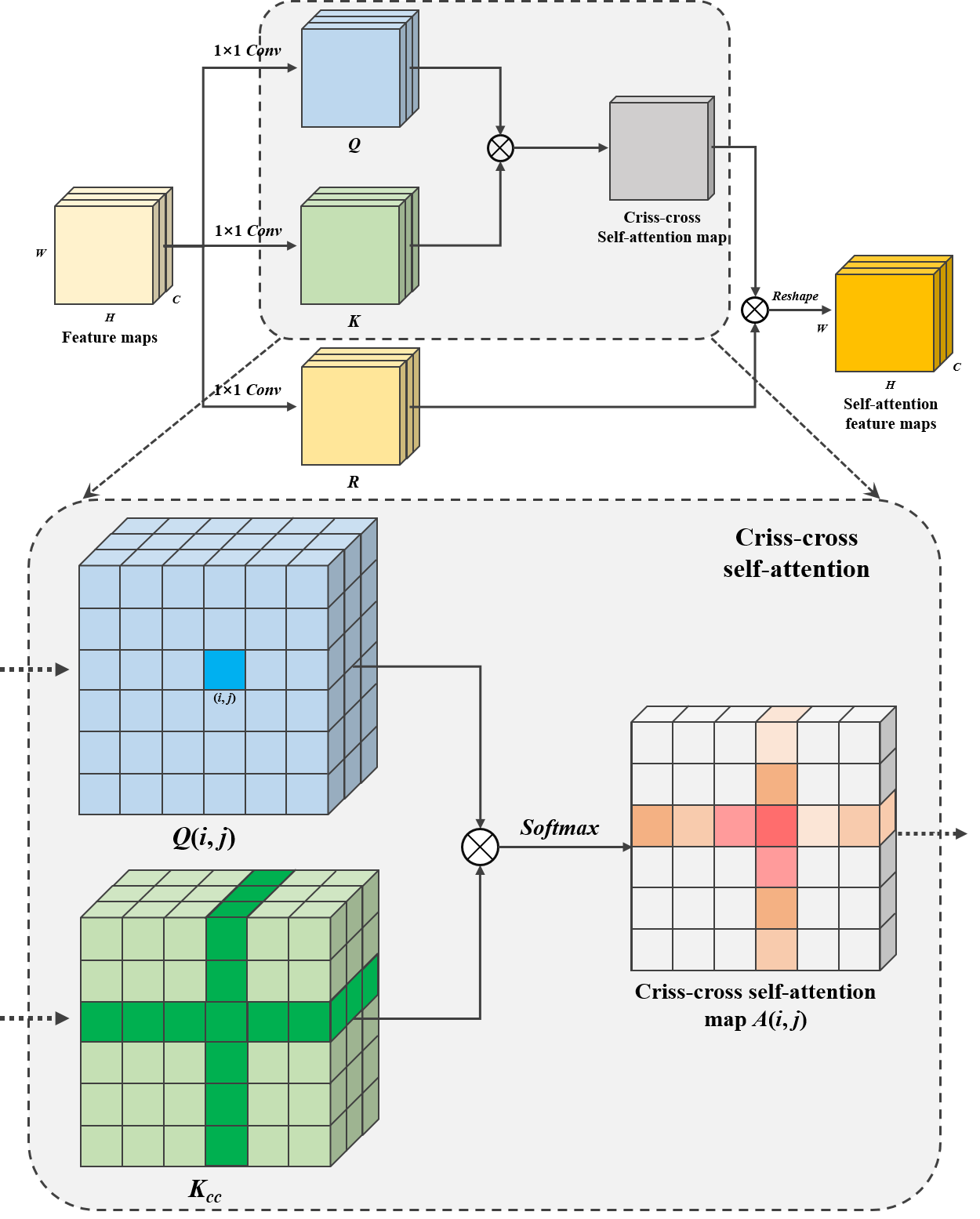}
  \label{fig:CSCD}}
  \caption{The Illustrations of (a) non-local mutual improvement, (b) recurrent information propagation, and (c) criss-cross self-attention module.}
  \label{fig:6}
\end{figure*}

\subsubsection{Criss-cross Self-attention Module for Lightweight Decoder}
\par As many studies have shown, highly discriminative features can help the classifier to achieve more precise change detection results \citep{Zhang2020, Shi2020, Gong2016}. Zhang \emph{et al.} \citep{Zhang2020} introduced a spatial attention module in the decoder network, which uses a large convolution kernel to fuse the local information for each position, thereby generating an attention map to separate changed and unchanged regions. However, this spatial attention module only utilizes the information in a local neighborhood for each position, but features in non-local positions can also provide information useful for highlighting feature representation. As shown in Fig. 6-(a), for a certain position belonging to the top emerged building, not only can features around this position help to improve discriminability, but features belonging to other emerged buildings can also offer information to aid in further distinguishing this pixel from unchanged ones. In turn, the feature representation in this position can help to improve the feature discriminability in the other changed buildings. The same is true for the unchanged pixels (see the blue illustration in Fig. 6-(a)). In order to utilize this non-local information to increase feature discriminability, we introduce a self-attention mechanism to model the relationship between a certain position and all other positions, thus ensuring that any two positions with similar features will contribute mutual discriminability regardless of their distance. For more information about the self-attention mechanism, please refer to \citep{Han2019}. 

\par Nonetheless, in the standard self-attention module, the time and space complexity of calculating a self-attention map are both $O(H W \times H W)$ \citep{Han2019}, which would result in high computational complexity and occupy a huge amount of GPU memory when processing large-scale VHR images. Accordingly, in EffCDNet, a criss-cross self-attention (CCA) mechanism \citep{Huang2019} is introduced to more efficiently model the dense relationship between pixels. Fig. 6-(c) presents the CCA module. Similar to the standard self-attention module, three feature maps $Q$, $K$, and $R$ are first generated from the input feature map by three 1$\times$1 convolutional layers. To optimize the self-attention map calculation step, for each pixel, the CCA module simply calculates the relationship between each pixel and the pixels in the horizontal and vertical directions rather than all remaining pixels, as follows:
\begin{equation}
D_{c c}(i, j)=Q(i, j) K_{c c}^{T}(i, j)
\label{eq:6}
\end{equation}
where $K_{c c}(i, j) \in R^{(H+W-1) \times C^{\prime}}$ is the set of features in the crisscross path of position $(i, j)$. A softmax operation is then applied on $D_{c c} \in R^{H \times W \times(H+W-1)}$ to calculate the attention map $A_{c c} \in R^{H \times W \times(H+W-1)}$. Finally, for the position $(i, j)$ in $V$, the set of features in the corresponding criss-cross path $V_{c c}(i, j) \in R^{(H+W-1) \times C}$ is collected and multiplied with $A_{c c}(i, j) \in R^{(H+W-1)}$ to obtain the corresponding feature representation in the self-attention feature map $H_{c c} \in R^{H \times W \times C}$ at position $(i, j)$: 
\begin{equation}
H_{c c}(i, j)=A_{c c}(i, j) V_{c c}(i, j)
\label{eq:7}
\end{equation}

\par Through employing this criss-cross calculation operation, the time and space complexity of the attention map calculation are greatly reduced from $O(H W \times H W)$ to $O(H W \times(H+W-1))$. However, a single CCA module can only capture information in the horizontal and vertical directions; the information in the remaining positions that are not in the criss-cross path is not considered. Accordingly, to harvest the dense information, the CCA module is duplicated twice in the network to form two loops. As shown in Fig. 6-(b), for the two positions $(x_{1}, y_{1})$ and $(x_{2}, y_{2})$ that are not in the same row and column, in the first loop, the information in the position $(x_{1}, y_{1})$ can propagate to position $(x_{2}, y_{1})$ and position $(x_{1}, y_{2})$. In the second loop, the information of $(x_{1}, y_{1})$ encoded in the position $(x_{2}, y_{1})$ and position $(x_{1}, y_{2})$ can be captured by position $(x_{2}, y_{2})$; thus, the information of position $(x_{1}, y_{1})$ can be indirectly obtained. Therefore, through the above recurrent CCA (RCCA) module, the similar features can be selectively integrated for each position to highlight their feature representations with a minor computational cost. Moreover, to further reduce parameters and computational cost, the three 1$\times$1 convolutional layers in the CCA module are replaced by the pointwise group convolution with channel shuffle.

\begin{figure*}[!t]
  \centering
  \subfloat[]{
    \includegraphics[width=1.05in]{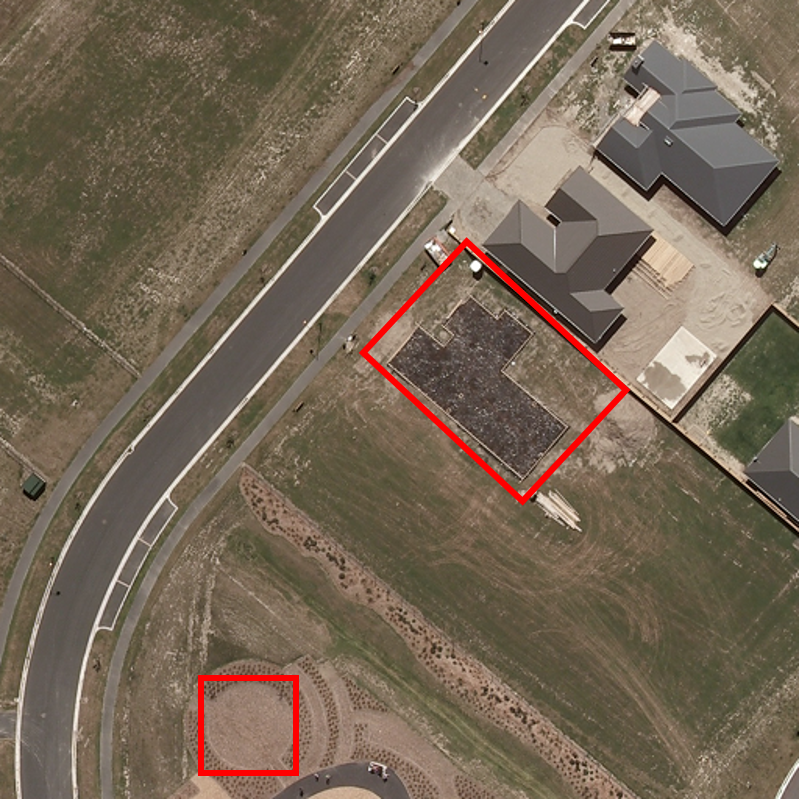}
  \label{fig:CSCD}}
  \hfil
  \subfloat[]{
    \includegraphics[width=1.05in]{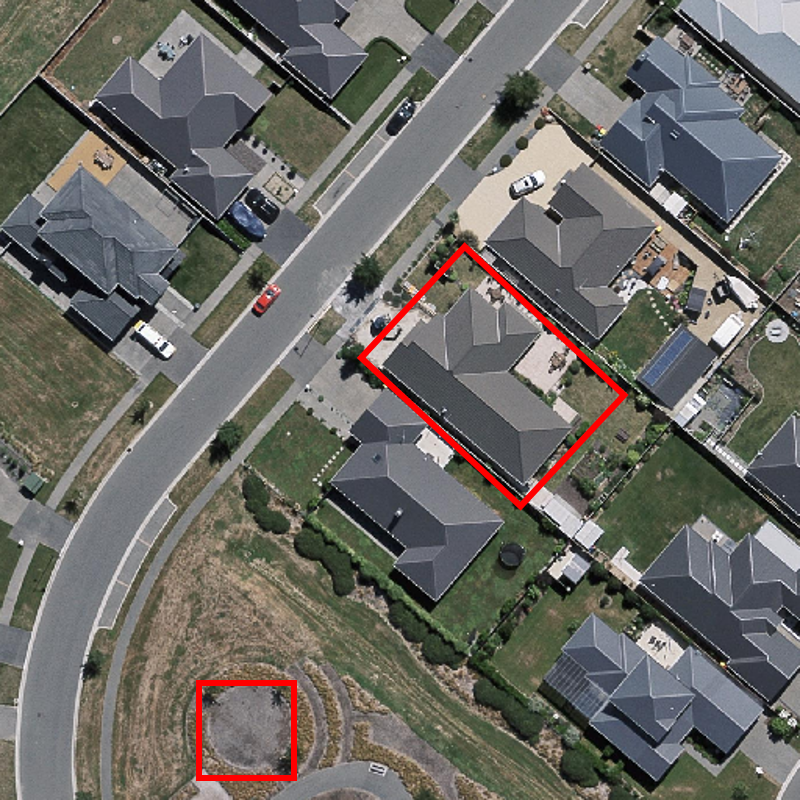}
  \label{fig:CSCD}}
  \hfil
  \subfloat[]{
    \includegraphics[width=1.05in]{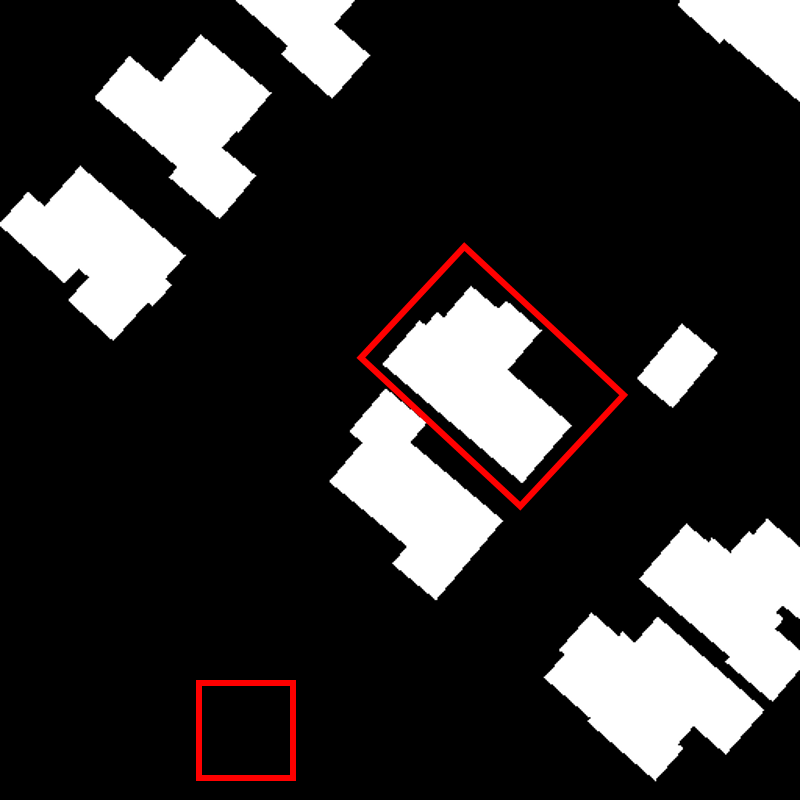}
  \label{fig:moti}}
  \hfil
  \subfloat[]{
    \includegraphics[width=1.05in]{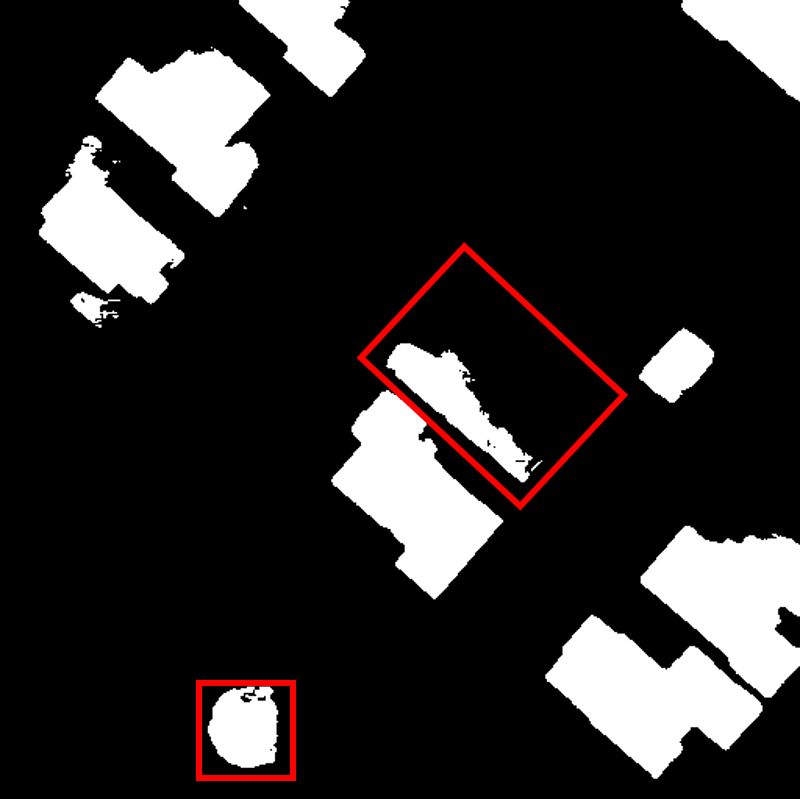}
  \label{fig:moti}}
  \hfil
  \subfloat[]{
    \includegraphics[width=1.05in]{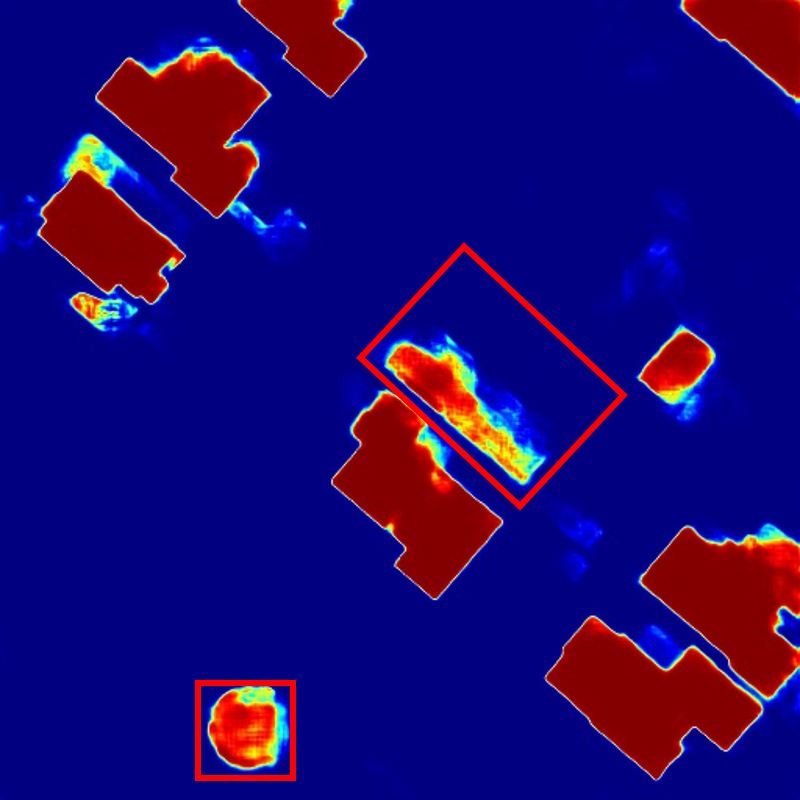}
  \label{fig:moti}}
  \caption{The illustration of confused pixels. (a) Pre-change image. (b) Post-change image. (c) Ground truth. (d) Change map generated by deep network. (e) Change probability map.}
  \label{fig:7}
\end{figure*}

\begin{figure}[t]
  \centering
  \includegraphics[scale=0.35]{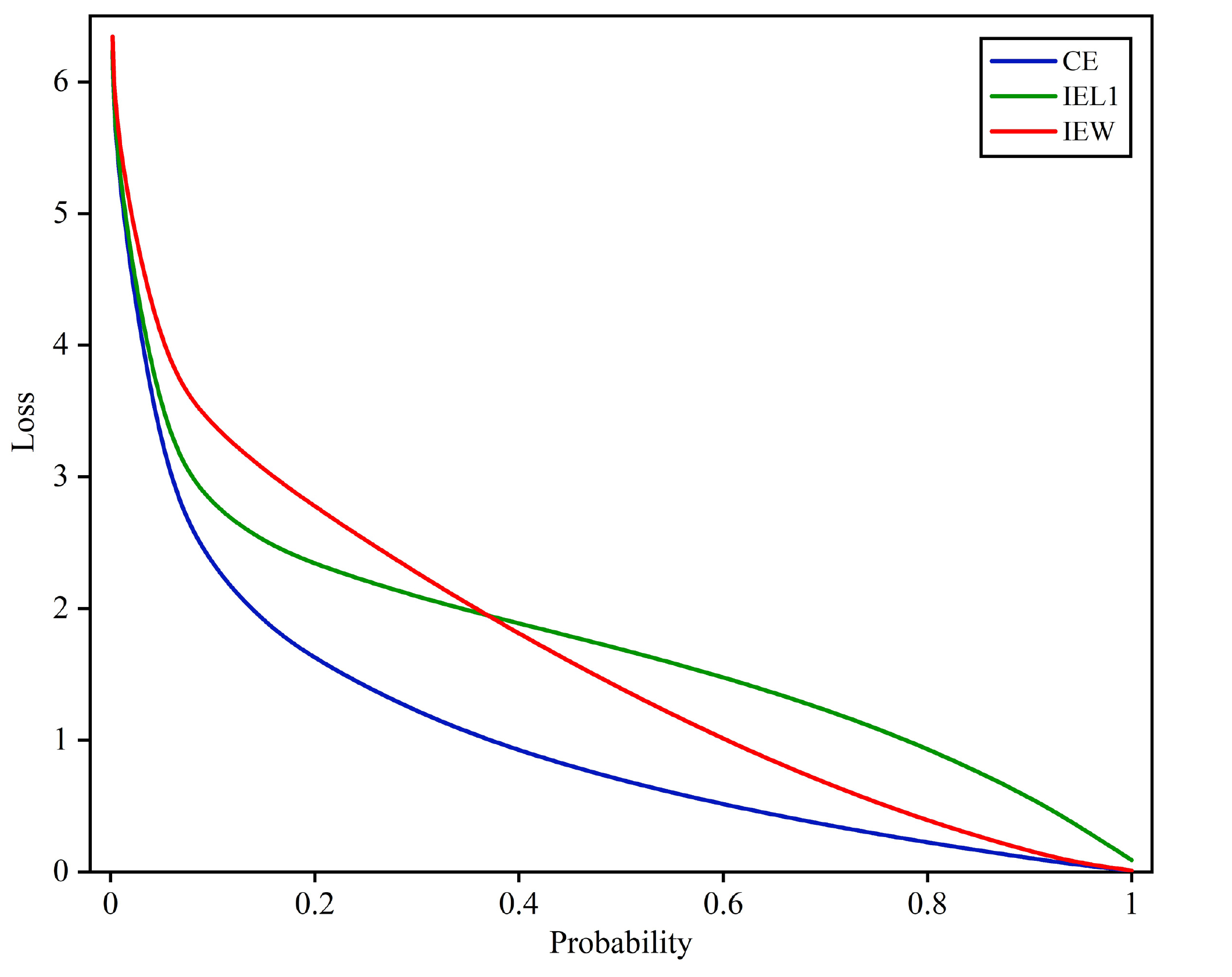}
  \caption{Curve of cross-entropy loss function and two information entropy-based loss functions.}
  \label{fig:8}
\end{figure}

\subsubsection{Shannon Information Entropy Loss for Network Training}
\par For change detection network training, cross-entropy loss is a commonly used loss function. However, as noted above, there exist some confused pixels that are difficult for cross-entropy loss to optimize (such as some non-building changes in building change detection tasks or pseudo-changes caused by seasonal variation). For these pixels, we argue that as the training stage progresses, the cross-entropy penalty is not large enough (see the middle interval of the blue line in Fig. \ref{fig:8}), and considering the ambiguity of confused pixels in change and non-change, the further optimization becomes difficult. Therefore, these pixels often have a similar probability of change and non-change in the prediction map. Fig. \ref{fig:7} illustrates the confused pixels in the building change detection task. As shown in Fig. 7-(e), most pixels have a large changed or unchanged probability (i.e., a very small changed probability), but the change probability of some confused pixels (marked with a red rectangle) is only about 0.5. 

\par Since these confused pixels have a close probability of change and non-change, from an information theory perspective, they are under-confident and have relatively large uncertainty. Shannon information entropy \citep{shannon2001mathematical} thus can be utilized to measure this uncertainty. Given a network prediction map $P \in R^{H \times W \times C}$, the entropy $E(i, j) \in[0,1]$ at position $(i, j)$ can be calculated as follows:
\begin{equation}
  E(i, j)=-\frac{1}{\log C} \sum_{c=1}^{C} P(i, j, c) \log P(i, j, c)
  \label{eq:8}
\end{equation}
where $C$ is the number of output categories, in change detection tasks, $C=2$. 

\par Based on information entropy, two novel loss functions are presented to help the cross-entropy loss in optimizing these confused pixels. In the entropy map of an ideal network output, only the pixels near the boundary between changed and unchanged regions have high entropy; the entropy of pixels within the region should be close to 0. With this in mind, we propose the first loss function, called information entropy L1 (IEL1) loss:
\begin{equation}
  L=\sum_{i=1}^{H} \sum_{j=1}^{W}\left(\sum_{c=1}^{C} Y(i, j) \log P(i, j)+\alpha\|E(i, j)-B(i, j)\|_{1}\right)
  \label{eq:9}
\end{equation}
Here, the first term is cross-entropy loss; in the second term, $B$ is an edge map produced from the ground truth by an edge detection algorithm. The second term uses $L_{1}$ loss to make the entropy map close to the ideal entropy map, thereby forcefully causing confused pixels with high entropy to have low entropy. Meanwhile, the cross-entropy loss would apply a large penalty to the wrong pixels to guarantee that their probabilities move towards the correct category. 

\par Moreover, considering that confused pixels have a relatively low penalty in cross-entropy loss but a fairly high entropy value, the second function, called information entropy weighted (IEW) loss, uses entropy to weigh each pixel, as follows:
\begin{equation}
L=\sum_{i=1}^{H} \sum_{j=1}^{W} \sum_{c=1}^{C}(1+E(i, j)) Y(i, j) \log P(i, j)
\label{eq:10}
\end{equation}
Through IEW loss, the confused pixels with high uncertainty can contribute to larger loss values, thus prompting the network to concentrate more on the optimization of confused pixels. 

\par Fig. \ref{fig:8} plots the curve of the cross-entropy loss function and two information entropy-based loss functions. We can therefore see that, compared with cross-entropy, IEL1 and IEW have an obviously large penalty in the intermediate intervals. The effects of information entropy loss functions are discussed in section 3.4.2 and section 3.4.4.

\section{Experimental Results and Analysis}\label{sec:3}
\subsection{Description of Datasets}
\par To validate the effectiveness of our proposed network, we conduct detailed experiments on two large-scale open-source change detection datasets. The first one, released in \citep{lebedev2018change}, is a season-varying dataset (SVCD) acquired from Google Earth, including seven season-varying image-pairs with a size of 4725$\times$2200 pixels and four season-varying image-pairs with a size of 1900$\times$1000 pixels. The spatial resolution of these images varies from 0.03m to 1m per pixel. The authors cropped these 11 multi-temporal image-pairs into image patch-pairs with a size of 256$\times$256 pixels \citep{lebedev2018change}. Finally, the SVCD dataset contains 10000 image-pairs for training, 3000 image-pairs for validation, and 3000 image-pairs for testing. 

\par The second dataset is a challenging building change detection dataset (BCD) \citep{Ji2019a}, which covers an area that was affected by a 6.3-magnitude earthquake in February 2011 and rebuilt in the following years. This dataset consists of one multi-temporal aerial image-pair with a size of 32507$\times$15354 pixels. The spatial resolution of these images is 0.075m per pixel. Considering the huge size of the two images, the two images are cropped into 1827 nonoverlapping 512$\times$512 image-pairs. We used 1096 (60$\%$) image-pairs to form the training set, 365 (20$\%$) image-pairs to form the validation set, and 366 (20$\%$) image-pairs to form the test set.

\begin{table*}[t] 
  \scriptsize
  \renewcommand{\arraystretch}{1.1}
  \caption{Specific network structure and configuration of EffCDNet}
  \label{table:1}
  \centering
  \begin{tabular}{c c c c c c c}
    \hline
    \bfseries Part            & \bfseries Module	             & \bfseries Depth      & \bfseries Layer & \bfseries	Repeat & \bfseries Output shape	& \bfseries Configuration \\
    \hline\hline
    \multirow{14}{*}{Encoder} & \multirow{3}{*}{Conv Block}	   & \multirow{3}{*}{3}   & Conv          & 1 & $H\times W\times 48$         & 3$\times$3, ReLU, BN \\ 											
    ~                         & ~	                             &  ~	                  & Conv	        & 2 & $H\times W\times 48$         & 3$\times$3, ReLU, BN \\ 												
    ~                         & ~	                             &  ~	                  & Max pooling	  & 1 & $H/2\times W/2\times 48$     & 3$\times$3, Stride 2 \\ 	
    \cline{2-7}	
    ~                         & \multirow{3}{*}{RCS Block I}   & \multirow{3}{*}{12}   & RCS unit	    & 1 & $H/2\times W/2\times 240$    & - \\ 	
    ~                         & ~	                             &  ~	                  & RCS unit	    & 2 & $H/2\times W/2\times 240$    & - \\ 	
    ~                         & ~	                             &  ~	                  & RCS unit	    & 1 & $H/4\times W/4\times 240$    & Stride 2 \\ 	
    \cline{2-7}
    ~                         & \multirow{3}{*}{RCS Block II}  & \multirow{3}{*}{75}  & RCS unit	    & 1 & $H/4\times W/4\times 480$    & - \\ 	
    ~                         & ~	                             &  ~	                  & RCS unit	    & 23 & $H/4\times W/4\times 480$    & - \\ 	
    ~                         & ~	                             &  ~	                  & RCS unit	    & 1 & $H/8\times W/8\times 480$    & Stride 2 \\ 	
    \cline{2-7}
    ~                         & \multirow{3}{*}{RCS Block III} & \multirow{3}{*}{12}  & RCS unit	    & 1 & $H/8\times W/8\times 960$    & - \\ 	
    ~                         & ~	                             &  ~	                  & RCS unit	    & 2 & $H/8\times W/8\times 960$    & - \\ 	
    ~                         & ~	                             &  ~	                  & RCS unit	    & 1 & $H/16\times W/16\times 960$  & Stride 2 \\ 	
    \cline{2-7}
    ~                         & Diff	                         & -	                  & Diff	        & 1 & $H/16\times W/16\times 960$  & - \\ 
    \cline{2-7}
    ~                         & EASPP	                         & 4	                  & see Fig. 5	  & 1 & $H/16\times W/16\times 256$  & - \\ 
    \hline
    \multirow{9}{*}{Decoder} & Upsample	                     & -                    & 4$\times$Upsample & 1 & $H/4\times W/4\times 256$ & Bilinear \\ 
    \cline{2-7}
    ~                         & \multirow{4}{*}{Skip connection} & \multirow{4}{*}{1} & Conv & 1 & $H/4\times W/4\times 48$ & 1$\times$1, ReLU, BN \\ 
    ~                         & ~                                 & ~                 & Conv & 1 & $H/4\times W/4\times 48$ & 1$\times$1, ReLU, BN \\ 
    ~                         & ~                                 & ~                 & Diff & 1 & $H/4\times W/4\times 48$ & - \\
    \cline{4-7}
    ~                         & ~                                 & ~                 & Concat & 1 & $H/4\times W/4\times 304$ & - \\ 
    \cline{2-7}			
    ~                         & Fusion                            & 1                 & EffConv & 1 & $H/4\times W/4\times 256$ & 1$\times$1 \\ 
    \cline{2-7}
    ~                         & RCCA                            & 2                 & CCA (see Fig. 6-(c)) & 2 & $H/4\times W/4\times 256$ & - \\ 
    \cline{2-7}
    ~                         & \multirow{3}{*}{Classifier}     & \multirow{3}{*}{7} & EffConv & 2 & $H/4\times W/4\times 256$ & 3$\times$3 \\ 
    ~                         & ~                               & ~                  & Conv   & 1 & $H/4\times W/4\times 2$ & 1$\times$1, Softmax \\ 
    ~                         & ~                               & ~                  & 4$\times$Upsample & 1 & $H\times W\times 2$ & Bilinear \\ 
    \hline
  \end{tabular}
\end{table*}

\subsection{Experimental Setup}
\subsubsection{Parameters Settings}
\par We implement our network based on Pytorch. The specific network structure and configuration of EffCDNet are listed in Table \ref{table:1}. It can be seen from the table that the encoder of EffCDNet is very deep and achieves a depth of 106 layers. By contrast, the decoder of EffCDNet is relatively shallow, with only has 10 layers. The depth of the entire network reaches 116 layers. To train the network, we apply a stochastic gradient descent (SGD) optimizer with an initial learning rate of 1$e^{-3}$, momentum of 0.9, and a weight decay of 1$e^{-6}$. The number of training epochs is set to 200. After the 100th epoch, the learning rate is halved if the validation loss does not drop at 10 epochs. The batch sizes are 16 and 4 for the SVCD and BCD datasets, respectively. Note that, on both datasets, the proposed EffCDNet is trained from scratch without the use of any pre-training technique. For the group numbers $G$ of group point-wise convolution, as a trade-off between accuracy and model size, we set $G=4$. For the recurrence of RCCA module $R$, since the dense relationship of each pixel to all remaining pixels can be captured when the loop number $R$ is 2, and considering the performance and computational complexity comprehensively, $R$ is set to 2. For the IEL1, we use the Canny edge detector \citep{Canny1986} on the ground truth to generate edge maps and set $\alpha$=1 as the balance between pixel category correctness and pixel certainty. Finally, on both datasets, we train the network with two information entropy-based loss functions and select the better of the two to generate the final change detection results. 

\par All experiments are conducted on an Intel Xeon E5-2620 v4 2.10-GHz processor and a single NVIDIA GTX 1080Ti GPU. The implementation of our method will be released through GitHub \footnote{https://github.com/I-Hope-Peace/EffCDNet}. 

\subsubsection{Evaluation Criteria}
\par To access the accuracy of the obtained change maps, four commonly used evaluation criteria are adopted: namely, precision rate (P), recall rate (R), overall accuracy (OA), and F1 score. P denotes the percentage of changed pixels that are correctly classified. R indicates the percentage of real changed pixels in all pixels classified into the change category. OA represents the number of both changed and unchanged pixels that are classified correctly, divided by the number of all pixels. Finally, F1 score is a comprehensive measure that considers both P and R, defined as the harmonic mean of P and R: $F_{1}=2 P R/(P+R)$. 

\par Furthermore, the parameter numbers of each model and the corresponding number of floating-point multiplication-adds (FLOPs) are utilized to quantitatively measure the complexity of different models. 

\begin{figure}[!t]
  \centering
  \subfloat[]{
    \includegraphics[width=0.50in]{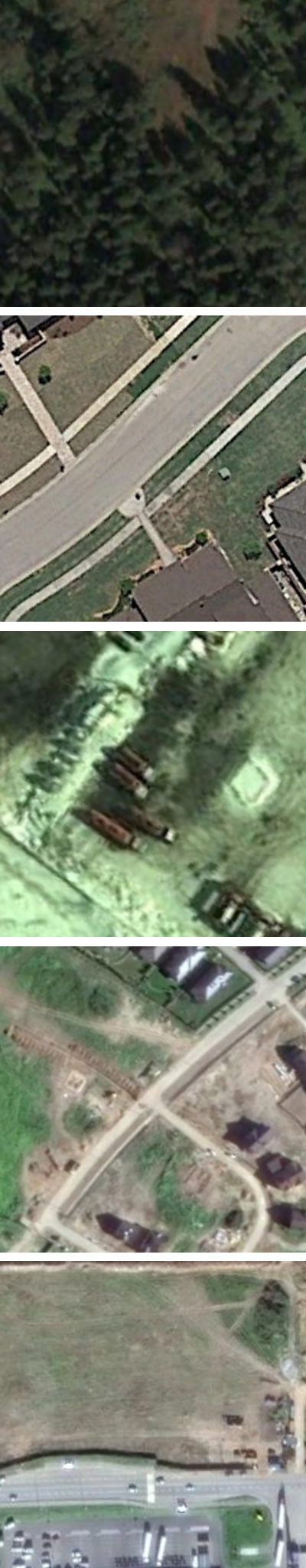}
  \label{fig:CSCD}}
  \centering
  \subfloat[]{
    \includegraphics[width=0.50in]{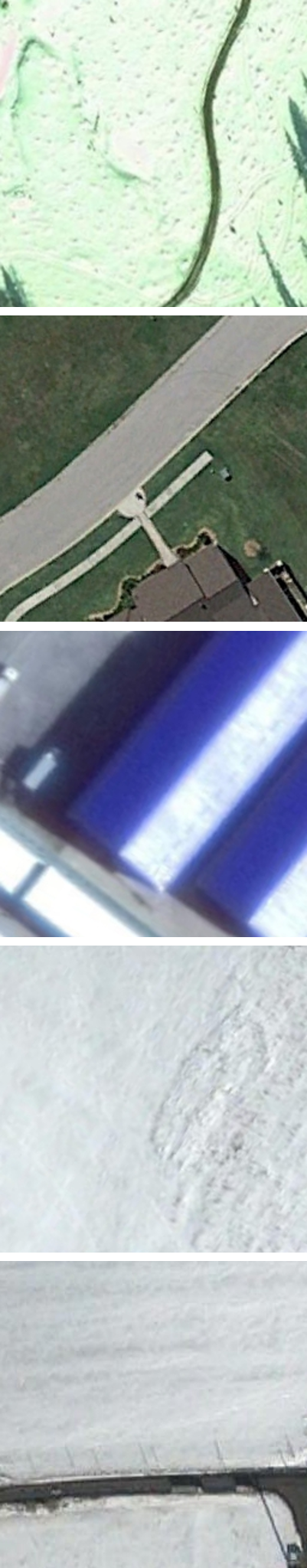}
  \label{fig:CSCD}}
  \centering
  \subfloat[]{
    \includegraphics[width=0.50in]{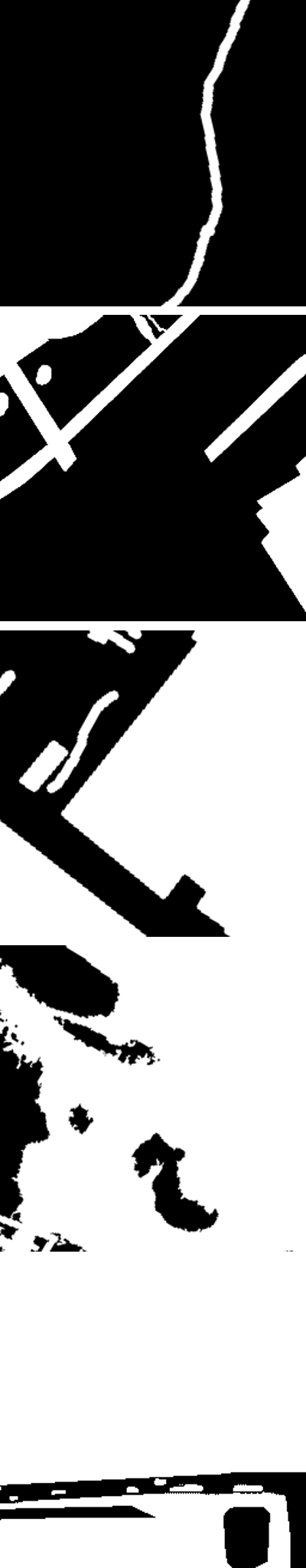}
  \label{fig:CSCD}}
  \centering
  \subfloat[]{
    \includegraphics[width=0.50in]{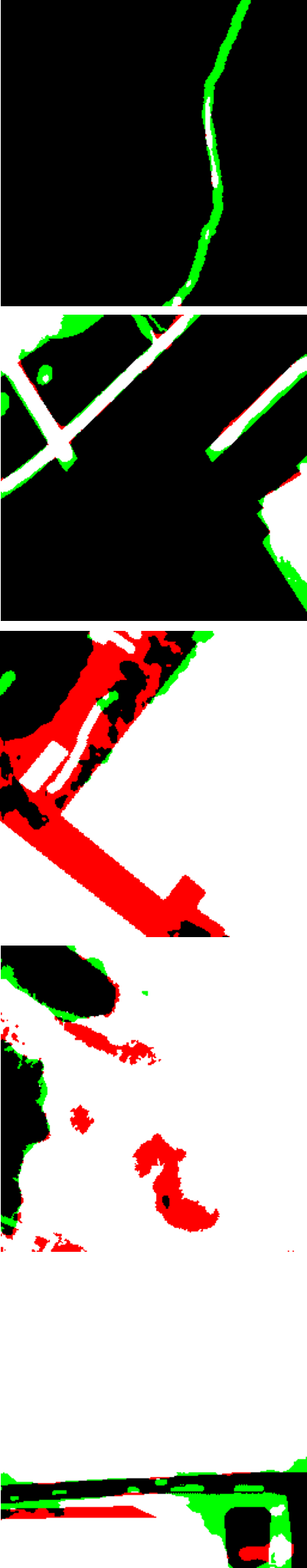}
  \label{fig:CSCD}}
  \centering
  \subfloat[]{
    \includegraphics[width=0.502in]{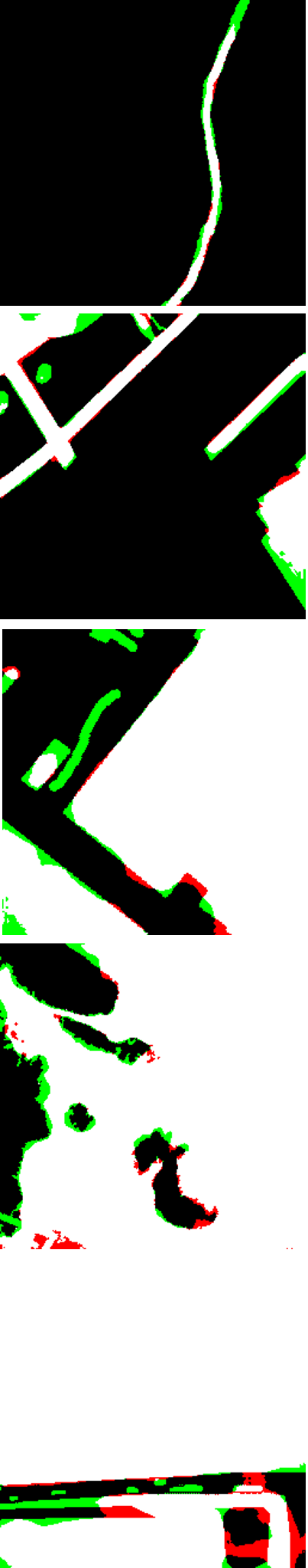}
  \label{fig:CSCD}}
  \centering
  \subfloat[]{
    \includegraphics[width=0.498in]{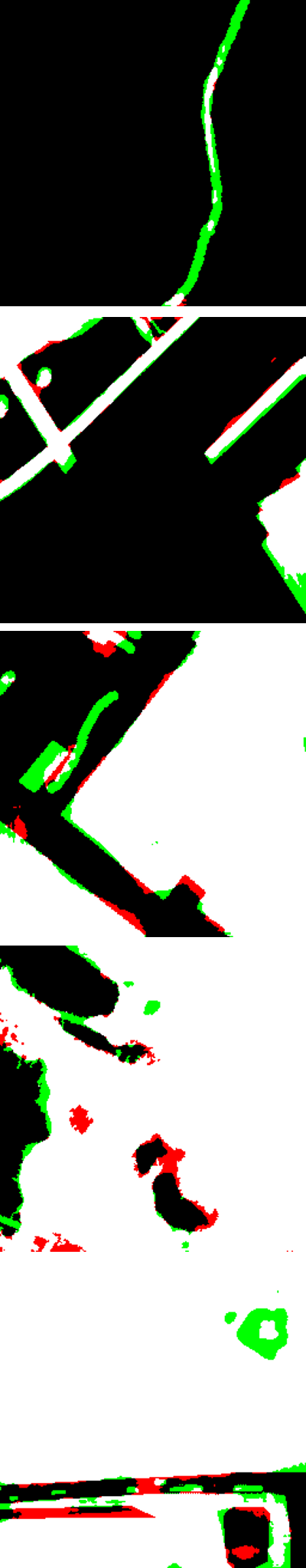}
  \label{fig:CSCD}}
  \centering
  \subfloat[]{
    \includegraphics[width=0.50in]{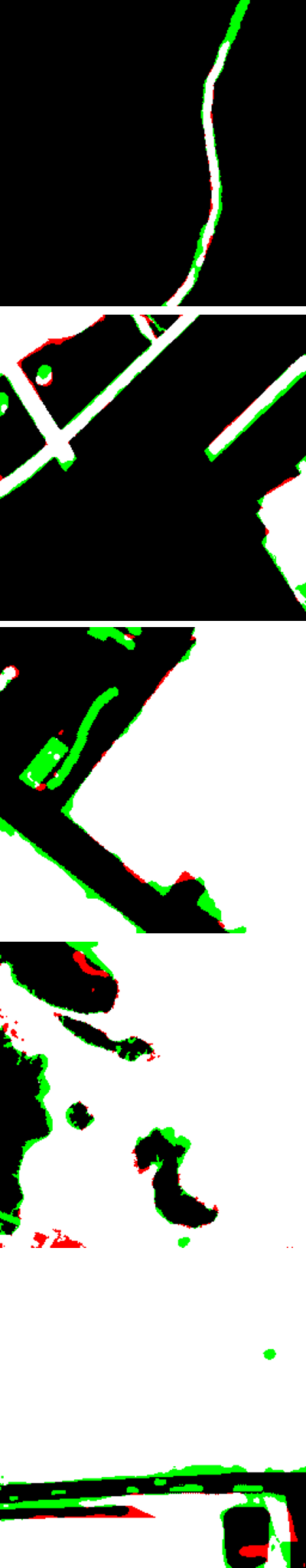}
  \label{fig:CSCD}}
  \centering
  \subfloat[]{
    \includegraphics[width=0.50in]{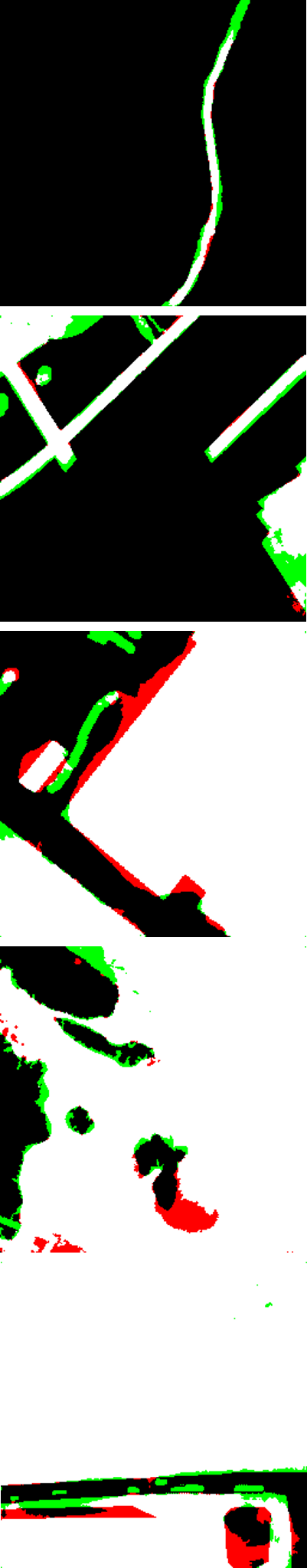}
  \label{fig:CSCD}}
  \centering
  \subfloat[]{
    \includegraphics[width=0.50in]{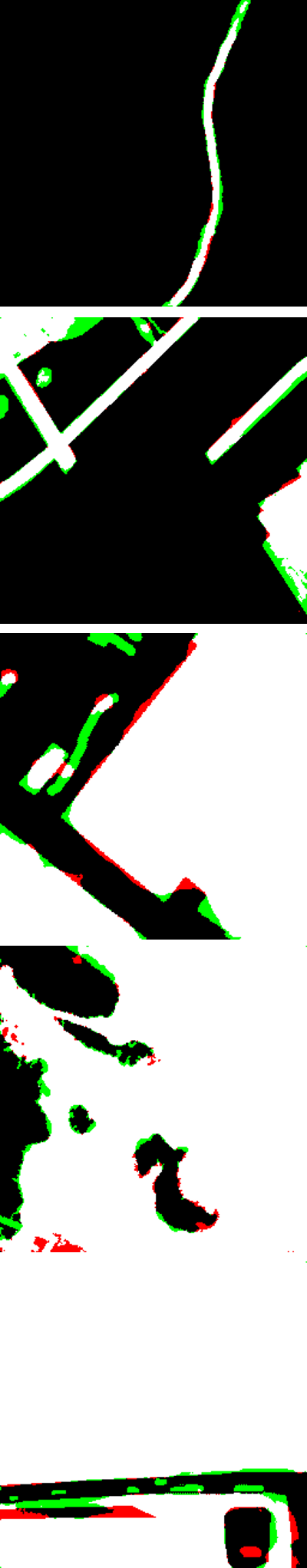}
  \label{fig:CSCD}}
  \centering
  \subfloat[]{
    \includegraphics[width=0.505in]{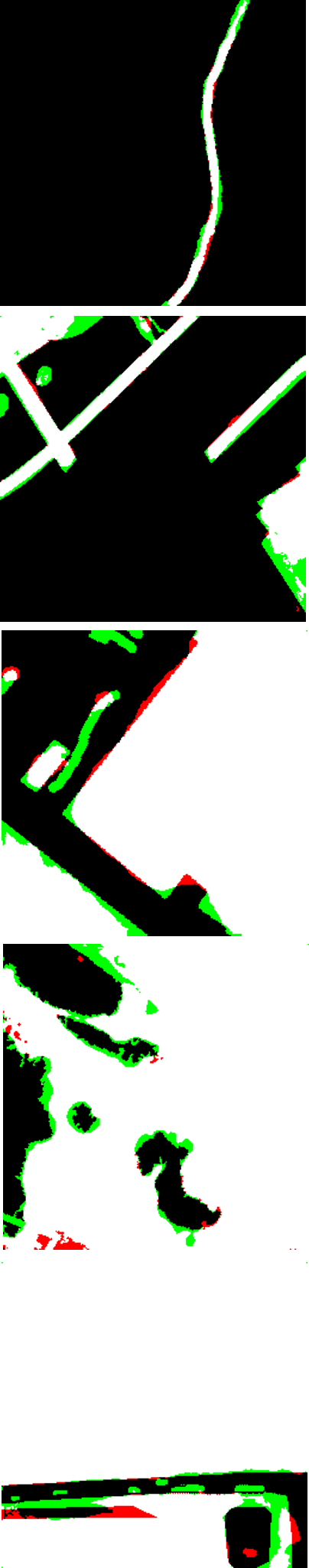}
  \label{fig:CSCD}}
  \centering
  \subfloat[]{
    \includegraphics[width=0.50in]{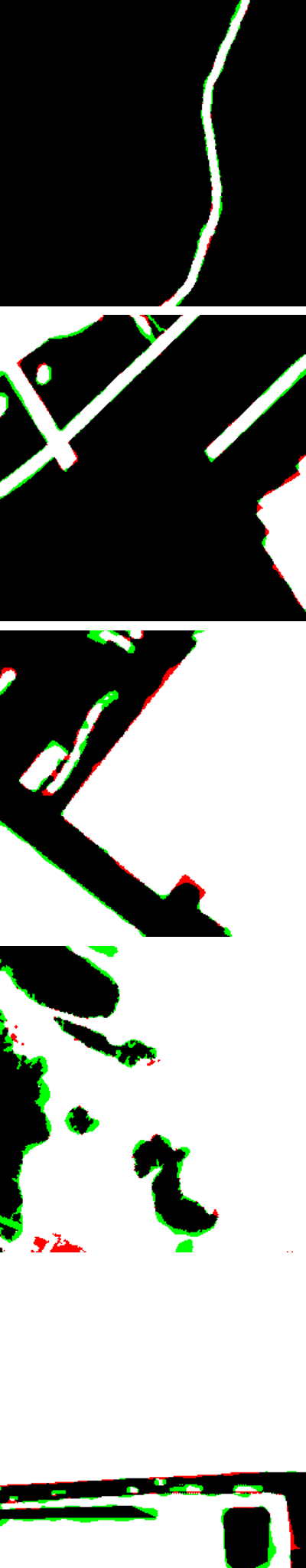}
  \label{fig:CSCD}}
  \caption{Change detection results obtained by different methods on the SVCD dataset. (a) Pre-change images. (b) Post-change images. (c) Ground truth. Change maps produced by: (d) FC-EF, (e) FC-Siam-Diff, (f) DSMS-FCN, (g) FCN-PP, (h) IUNet++, (i) DASNet, (j) IFN, and (k) EffCDNet. In change maps, white areas indicate true changed pixels, black areas indicate true unchanged pixels, red areas represent unchanged pixels that are falsely detected as changed pixels (false alarm), and green areas represent changed pixels that are falsely detected as unchanged pixels (missing alarm).}
  \label{fig:9}
\end{figure}

\subsection{Change Detection Results}

\par To validate the effectiveness and superiority of our proposed EffCDNet, we conduct experiments on the two datasets and compare EffCDNet with the seven comparison methods, including FC-EF \citep{CayeDaudt2018}, FC-Siam-Diff \citep{CayeDaudt2018}, DSMS-FCN \citep{Chen2019}, FCN-PP \citep{Lei2019}, IUNet++ \citep{Peng2019a}, DASNet \citep{chen2020dasnet}, and IFN \citep{Zhang2020}. 

\par Fig. \ref{fig:9} presents some change detection maps produced by our EffCDNet and seven comparison methods on the SVCD dataset. In the five illustrated examples, the types of change are various, the scales of the changed regions also differ, and there are many interferences caused by seasonal variation. As shown in this figure, all change maps acquired by our model are very close to the ground truth, the changed regions are complete and accurate, and pseudo-changes are well suppressed. For example, the changes in the example of the first row show the emergence of a road. Although the road is very narrow and the changes in vegetation caused by seasonal variation are obvious, the proposed method still successfully detects road changes without fragmentation and is not disturbed by the seasonal changes. In short, all visual interpretations relating to the change maps in Fig. \ref{fig:9} qualitatively reflect the effectiveness of the proposed EffCDNet. 

\par To confirm the superiority of EffCDNet, in Table \ref{table:2}, we report the quantitative results of the comparison models and EffCDNet on the SVCD dataset. First, we could see that the depth of EffCDNet is far deep than that of these comparison methods. As is evident, the proposed method outperforms all comparison methods on the four evaluation criteria. Limited to the simple network architecture, the accuracies of the two benchmark models (i.e., FC-EF and FC-Siam-Diff) are much lower than that of EffCDNet. Despite adopting a pyramid pooling module to extract multi-scale features, the early fusion strategy is not beneficial for highlighting change information, leading to the poor performance of FCN-PP. In \citep{Zhang2020}, by utilizing pretraining technology for the encoder, as well as deep supervision and attention module for the decoder, the proposed IFN achieves promising results on the SVCD dataset. Compared to IFN, the proposed EffCDNet gets a slightly better value of P and a considerable increase in terms of R, which means that the proposed EffCDNet is capable of detecting more complete changed regions. Finally, EffCDNet achieves improvements in terms of accuracy by 1.18$\%$ of OA, and 4.97$\%$ of F1, respectively, in comparison with IFN.

\begin{figure*}[!t]
  \centering
  \subfloat[]{
    \includegraphics[width=0.50in]{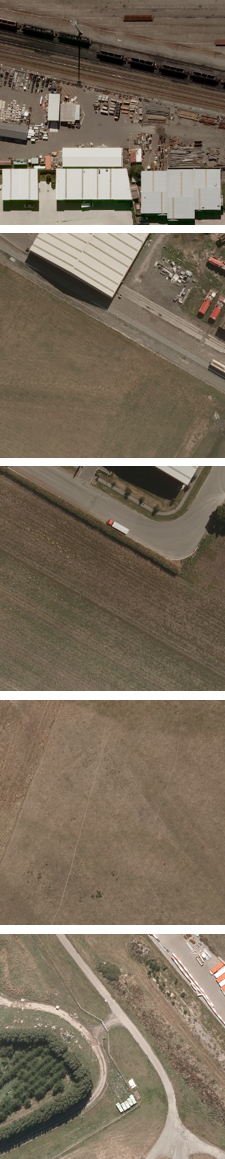}
  \label{fig:CSCD}}
  \subfloat[]{
    \includegraphics[width=0.50in]{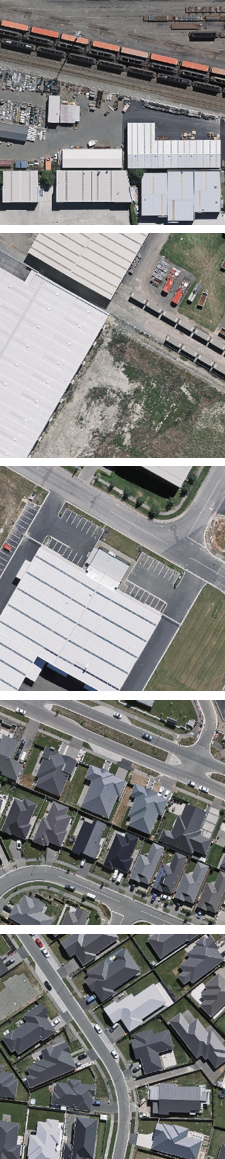}
  \label{fig:CSCD}}
  \subfloat[]{
    \includegraphics[width=0.50in]{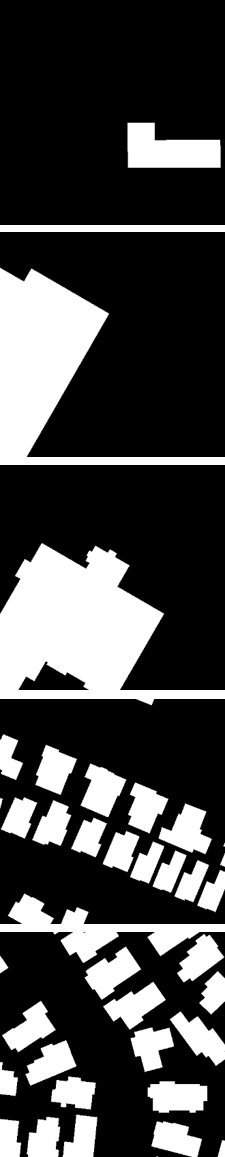}
  \label{fig:CSCD}}
  \subfloat[]{
    \includegraphics[width=0.50in]{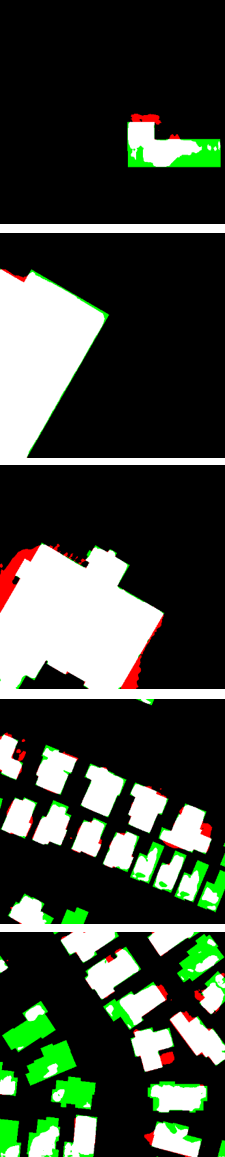}
  \label{fig:CSCD}}
  \subfloat[]{
    \includegraphics[width=0.50in]{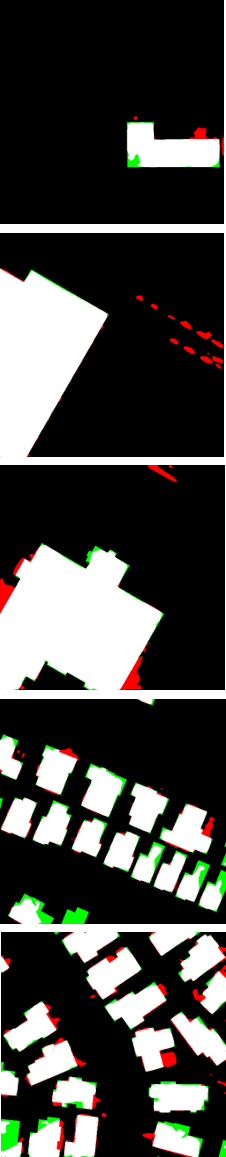}
  \label{fig:CSCD}}
  \subfloat[]{
    \includegraphics[width=0.50in]{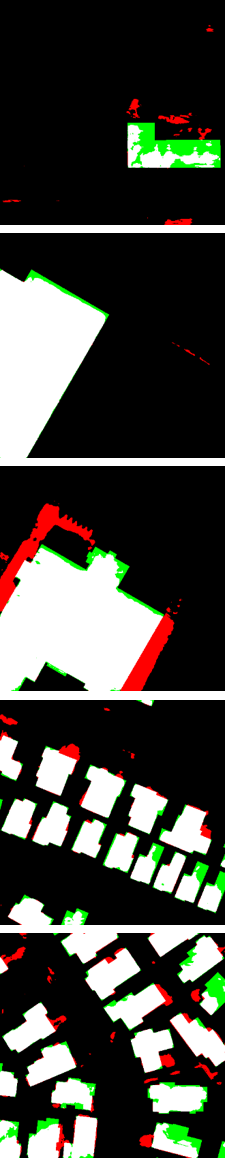}
  \label{fig:CSCD}}
  \subfloat[]{
    \includegraphics[width=0.50in]{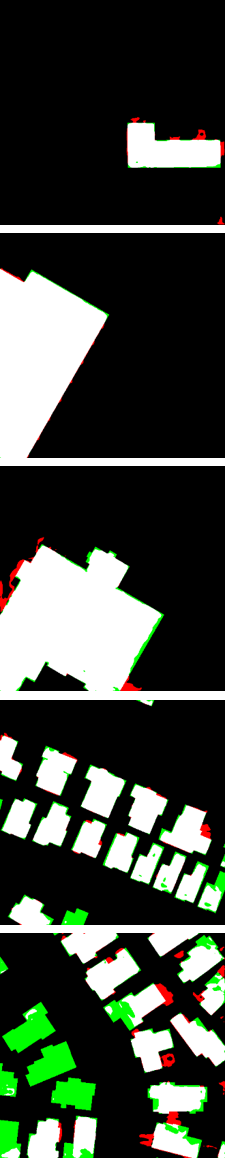}
  \label{fig:CSCD}}
  \subfloat[]{
    \includegraphics[width=0.50in]{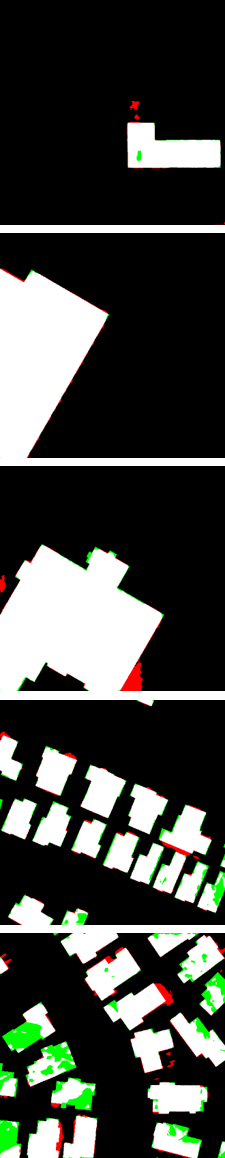}
  \label{fig:CSCD}}
  \subfloat[]{
    \includegraphics[width=0.50in]{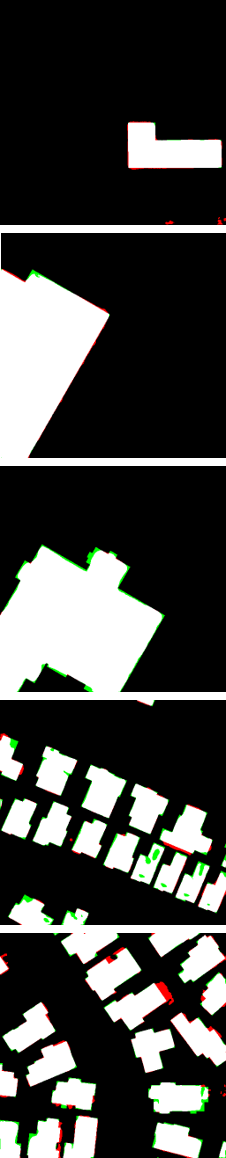}
  \label{fig:CSCD}}
  \subfloat[]{
    \includegraphics[width=0.50in]{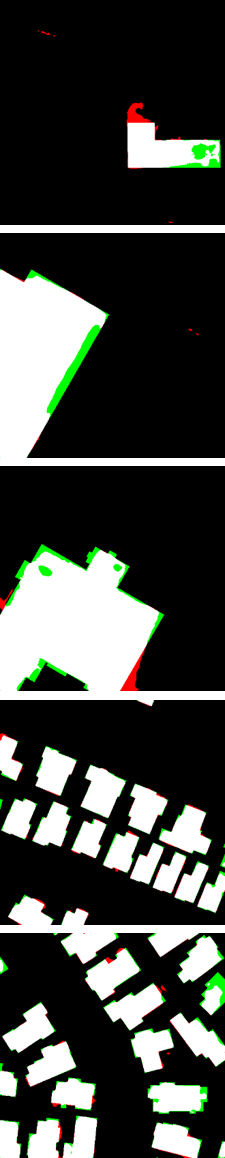}
  \label{fig:CSCD}}
  \subfloat[]{
    \includegraphics[width=0.50in]{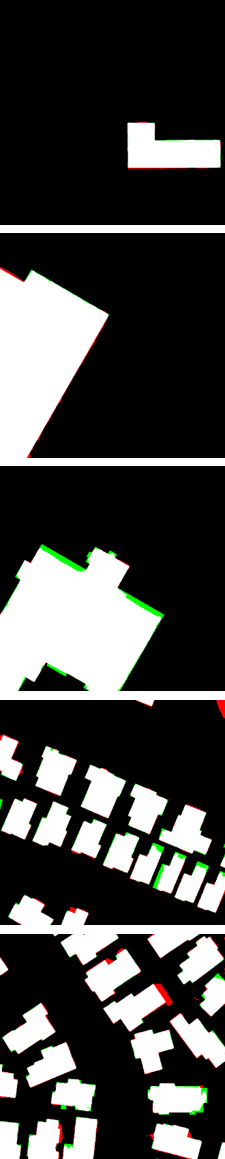}
  \label{fig:CSCD}}
  \caption{Building change detection results obtained by different methods on the BCD dataset. (a) Pre-change images. (b) Post-change images. (c) Ground truth. Change maps produced by: (d) FC-EF, (e) FC-Siam-Diff, (f) DSMS-FCN, (g) FCN-PP, (h) IUNet++, (i) DASNet, (j) IFN, and (k) EffCDNet. In change maps, white areas indicate true changed pixels, black areas indicate true unchanged pixels, red areas represent unchanged pixels that are falsely detected as changed pixels (false alarms) and green areas represent changed pixels that are falsely detected as unchanged pixels (missing alarm).}
  \label{fig:10}
\end{figure*}
		
\begin{table}[t]
  \scriptsize

  \renewcommand{\arraystretch}{1.2}
  \caption{Accuracy assessment of change detection results produced by various methods on the SVCD dataset} 
  \label{table:2}
  \centering
  \begin{tabular}{c c c c c c c}
    \hline
    \bfseries Method	& \bfseries P ($\%$) & \bfseries	R ($\%$) & \bfseries	OA ($\%$)	& \bfseries F1 ($\%$) &  \bfseries Depth \\
    \hline\hline
    FC-EF	& 81.00	& 73.34	& 94.82	& 77.00	 & 20 \\ 												
    FC-Siam-Diff	& 88.62	& 80.29	& 96.45	& 84.25	&  20 \\ 
    DSMS-FCN  & 89.34		& 82.40	& 	96.85 	& 85.73	& 28 \\ 								 							
    FCN-PP  & 91.77	& 82.21	& 97.03	 	& 86.73	 & 16 \\ 		
    IUNet++	& 89.54	& 87.11	& 96.73	& 87.56 & 24 \\  	  
    DASNet  & 92.52 & \underline{91.45}	& \underline{98.07} & \underline{91.93}	& 19 \\ 											
    IFN	& \underline{94.96}		& 86.08	& 97.71	& 90.30  & 37 \\ 	
    EffCDNet	&  \textbf{95.68}	& \textbf{94.86}	& \textbf{98.89}	& \textbf{95.27} & \textbf{116} \\ 	
    \hline
  \end{tabular}
\end{table}
\par To further demonstrate the generality of EffCDNet in change detection tasks, we conduct experiments on the BCD dataset. This dataset is a building change detection dataset, in which only building changes belong to the change class; other types of changes are pseudo-changes and should be classified into the non-change class. Fig. \ref{fig:10} presents some building change maps produced by our model and the other seven comparison methods. From the figure, we can see that EffCDNet generates the most accurate change maps out of all methods. For different types, numbers, and scales of building changes, EffCDNet can generate changed regions with precise boundaries and high internal compactness. For the other types of changes, regardless of their change intensity, EffCDNet is capable of correctly classifying them into the non-change class. Although the state-of-the-art method IFN can detect complete changed regions, the internal compactness of the obtained results is not high (see the first and third rows of Fig. 10-(i)), meaning that it cannot compete with EffCDNet.  
					
\begin{table}[t]
  \scriptsize

  \renewcommand{\arraystretch}{1.2}
  \caption{Accuracy assessment of change detection results produced by various methods on the BCD dataset}
  \label{table:3}
  \centering
  \begin{tabular}{c c c c c c c}
    \hline
    \bfseries Method	& \bfseries P ($\%$) & \bfseries	R ($\%$) & \bfseries	OA ($\%$)	& \bfseries F1 ($\%$) &  \bfseries Depth \\
    \hline\hline
    FC-EF	& 72.94	& 81.76		& 95.46		& 	77.10 & 20 \\ 																						
    FC-Siam-Diff	& 	80.58		& 77.47	& 96.15		& 79.00	&  20 \\ 
    DSMS-FCN  & 84.30		& 81.33	& 96.83	 	& 82.79&  28	 \\ 					
    FCN-PP  & 84.37		& 80.87	& 96.81	 	& 82.58	& 16 \\ 			
    IUNet++	& 90.88	& 81.41	& 97.50	& 85.89 & 24 \\  	
    DASNet  & \underline{90.94}	& 85.16	& \underline{97.82} 	& \underline{88.00} &  19 \\ 											
    IFN	& 89.10		& \underline{85.52}	& 97.67	& 87.27 & 37 \\ 	
    EffCDNet	&  \textbf{92.54}	& \textbf{90.06}	& \textbf{98.39}	& \textbf{91.29} & \textbf{116} \\ 	
    \hline
  \end{tabular}
\end{table}
\par Table \ref{table:3} presents the quantitative results on the BCD dataset. Once again, the proposed EffCDNet achieves the best performance on all evaluation criteria. By using a pretraining network, deep supervision technique, and attention module, IFN obtains decent performance. In comparison, through using a very deep encoder, an efficient multi-scale feature extraction module, and a light decoder with self-attention mechanism, our EffCDNet achieves obvious performance increases of 3.44$\%$, 4.54$\%$, 0.72$\%$, and 4.02$\%$ for P, R, OA, and F1, respectively.

\begin{figure}[!t]
  \centering
  \includegraphics[scale=0.4]{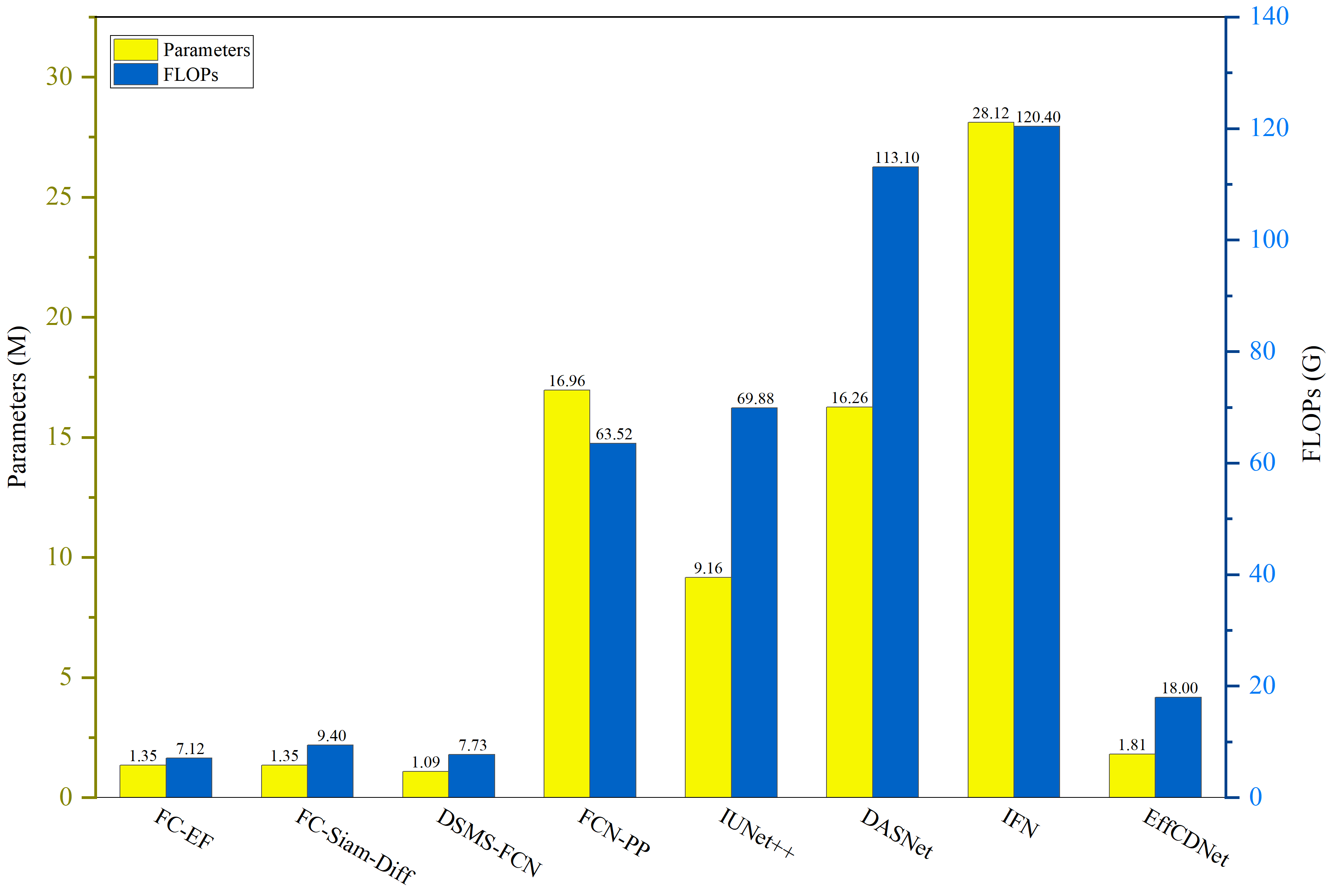}
  \caption{The parameter numbers and computational amount of EffCDNet and comparison methods. FLOPs are estimated under one bi-temporal image-pair with a size of 2$\times$3$\times$256$\times$256.}
  \label{fig:11}
\end{figure}

\par On both datasets, the proposed EffCDNet generates highly precise change maps and outperforms all other comparison methods on all evaluation criteria, which demonstrates the effectiveness and superiority of our model. Going one step further, Fig. \ref{fig:11} plots the parameter numbers of EffCDNet and comparison methods, along with their FLOPs, when processing a pair of bi-temporal images with a size of 3$\times$256$\times$256. From this figure, we can derive more promising conclusions. First, due to the shallow and simple network architecture, the two benchmark methods (i.e. FC-EF, FC-Siam-Diff) have very few parameters and a low computational cost; nevertheless, it is difficult to fully guarantee the accuracy of these methods. By contrast, EffCDNet achieves the best performance on both datasets with a network depth of 116 layers, but the model size of EffCDNet is almost the same as the benchmark methods, while its computational cost is only slightly higher than FC-Siam-Diff. Although IFN achieved very high accuracy in the SVCD dataset, its utilization of standard convolution in deep layers and the complicated decoder network result in too many parameters (28.12M) and very high computational cost (120.40 GFLOPs). In contrast, by adopting the efficient convolution to replace the standard convolution, the parameter numbers of EffCDNet is 6.4$\%$ of that of IFN and computational cost is only 14.9$\%$ of that of IFN. Moreover, the proposed EffCDNet also achieves better performance than IFN on both datasets. In summary, EffCDNet successfully balances the contradiction between change detection performance and the computational burden caused by deep architecture. Under the same hardware conditions, compared to the state-of-the-art methods, EffCDNet requires fewer computational resources but can achieve higher accuracy, which makes it superior overall in an actual production environment.

\subsection{Discussion}
\subsubsection{Network training}

\begin{figure*}[!t]
  \centering
  \subfloat[]{
    \includegraphics[scale=0.21]{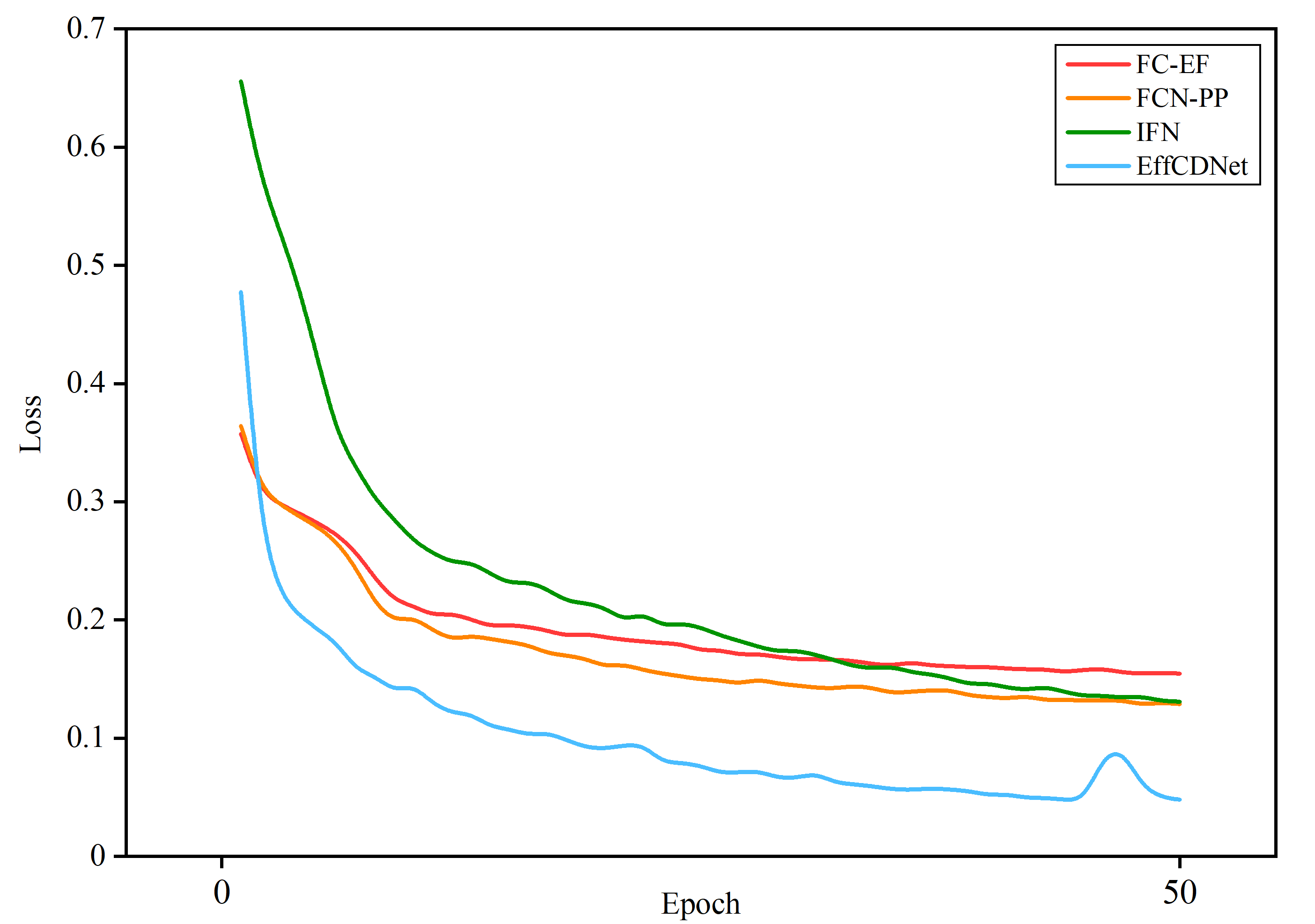}
  \label{fig:CSCD}}
  \hfil
  \subfloat[]{
    \includegraphics[scale=0.21]{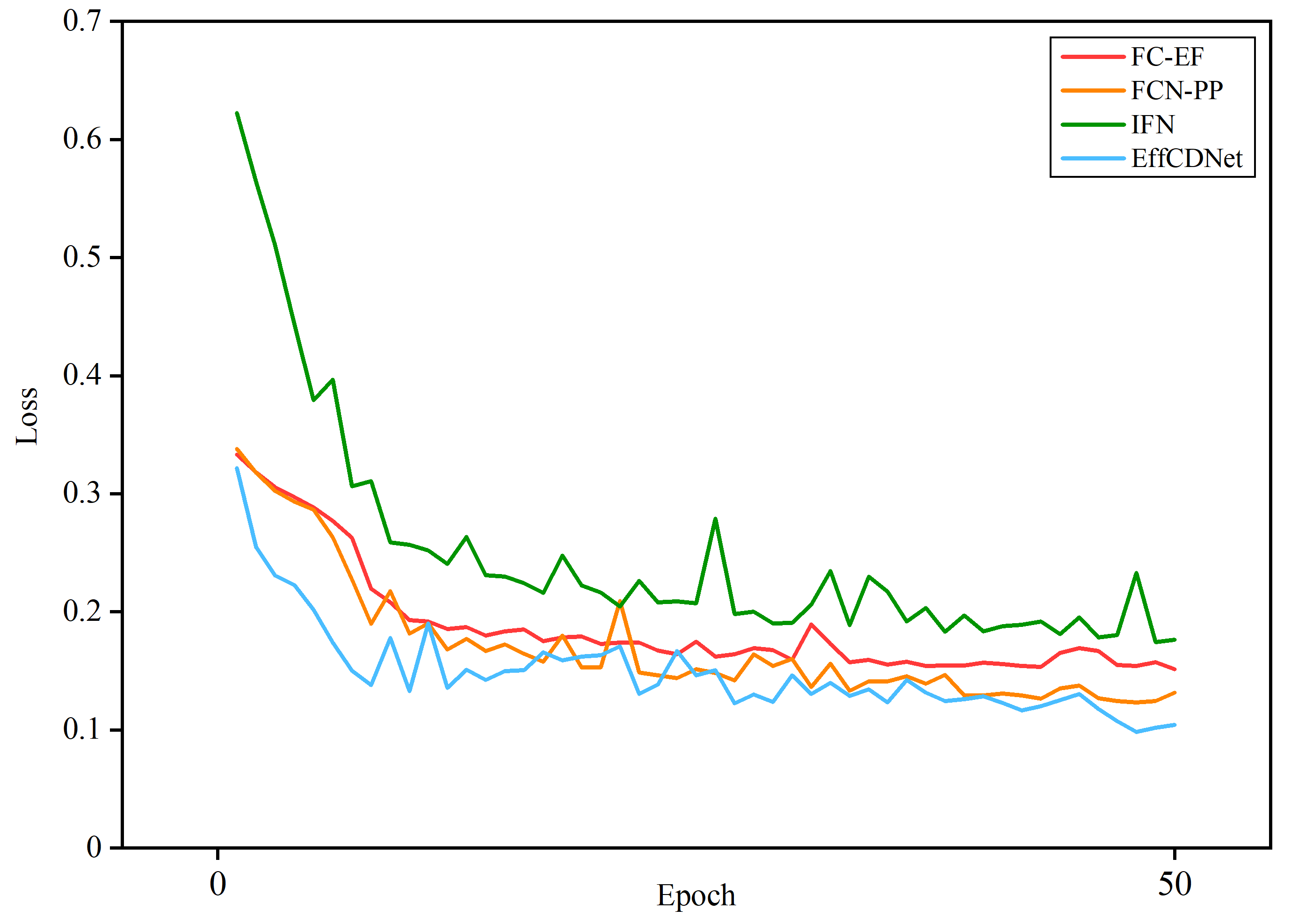}
  \label{fig:CSCD}}
  \hfil
  \subfloat[]{
    \includegraphics[scale=0.21]{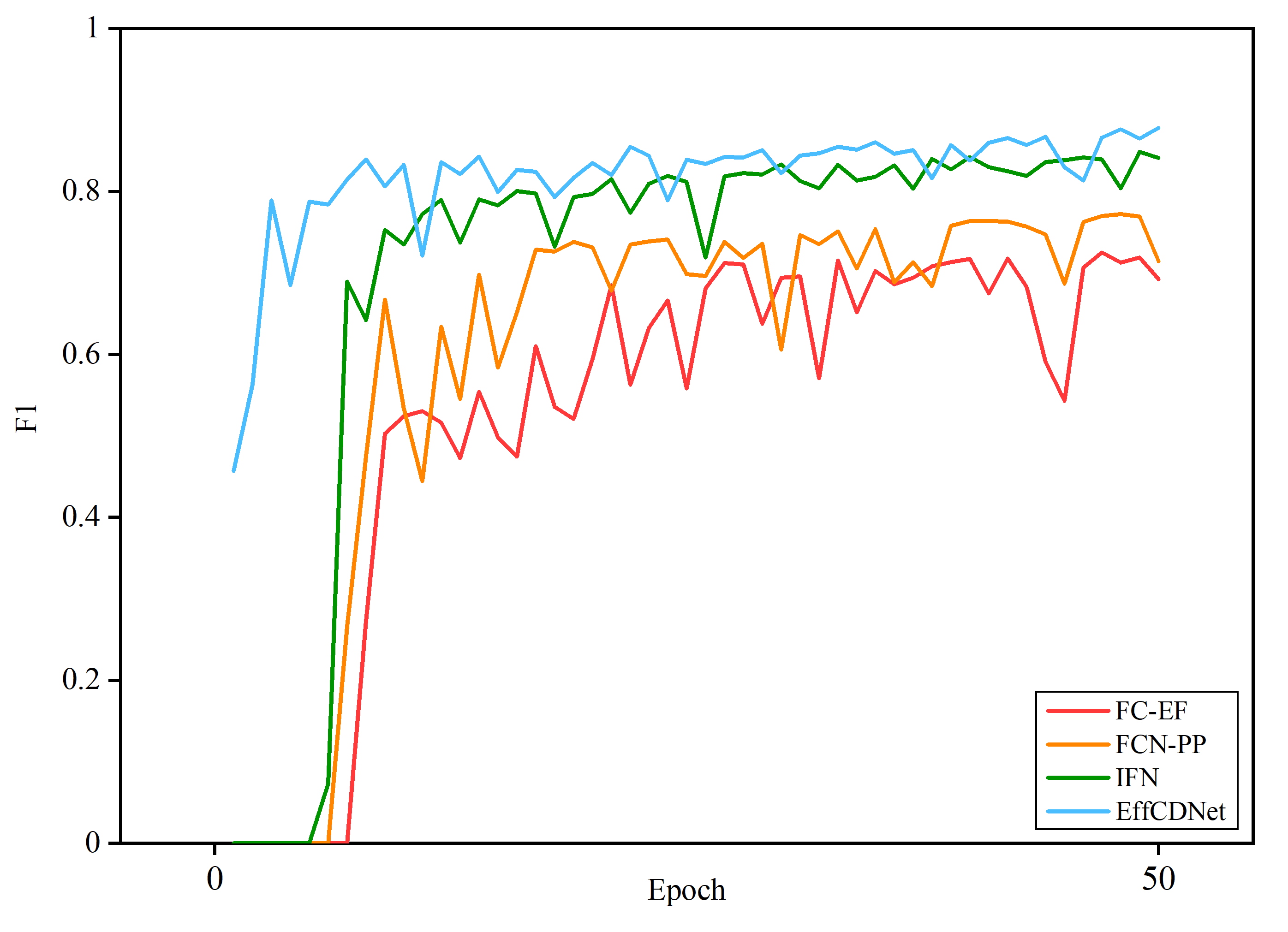}
  \label{fig:moti}}
  \caption{Learning curves for the four methods on the BCD dataset. (a) Training loss. (b) Validation loss. (c) Validation F1.}
  \label{fig:12}
\end{figure*}

\par Due to the training problems caused by deep architecture, existing change detection networks are not deep enough, which limits their learning ability. Moreover, some complex networks, such as IFN, use a pre-training network for parameter initialization and deep supervision techniques to alleviate training problems. Even so, the network depth of IFN is only 37 layers. In comparison, through its introduction of residual learning in the RCS unit, EffCDNet does not encounter training problems despite having a network depth of 116 layers. Fig. \ref{fig:12} plots the learning curves of the proposed network and three comparison networks (FC-EF, FCN-PP, IFN) trained on the BCD dataset. As shown in Fig. 12-(a), in the training set, compared to the other three methods, the loss of EffCDNet drops fastest and becomes the smallest after the first few epochs, which demonstrates the superiority of deep architecture with residual learning. Furthermore, EffCDNet exhibits very good generalization ability. As presented in Fig. 12-(b), in the validation set, the loss of EffCDNet gradually decreases and achieves the smallest value. Fig. 12-(c) further shows the F1 obtained by the four methods on the validation set. We can see that the F1 of EffCDNet reaches a value close to 0.5 after the first epoch and surpasses 0.8 at the seventh epoch. Subsequently, the F1 of EffCDNet continues to rise steadily until it reaches 0.88 at the 50th epoch. In comparison, through the use of facilitating training techniques, after the first few epochs, the F1 of IFN also rises quickly and then maintains a steady increase; however, it is still lower than that of EffCDNet. 

\begin{table}[!t]
  \scriptsize

  \renewcommand{\arraystretch}{1.1}
  \caption{Performance contribution of key components in EffCDNet on both datasets}
  \label{table:4}
  \centering
  \begin{tabular}{c c c c c c c}
    \hline
    \bfseries Dataset & \bfseries Method	 & \bfseries EASPP & \bfseries RCCA & \bfseries	IEL1 & \bfseries IEW	& \bfseries F1 ($\%$) \\
    \hline\hline
    \multirow{6}{*}{SVCD} & Backbone	&  	&  		&  		& 	  & 	91.73 \\ 											
    ~ & +EASPP	& $\checkmark$		& 		&  	&  	 & 	 93.72 \\ 												
    ~ & +RCCA	& 	 		& $\checkmark$	& 		&   & 94.30 \\ 					 							
    ~ & EffCDNet & $\checkmark$		& $\checkmark$	&   	&  	&  94.63 \\ 				
    ~ & +IEL1	& $\checkmark$		& $\checkmark$	& $\checkmark$	&   & 95.27\\ 	
    ~ & +IEW	&  $\checkmark$	& $\checkmark$	&  & $\checkmark$ & 95.16  \\ 	
    \hline
    \multirow{6}{*}{BCD} & Backbone	&  	&  		&  		& 	  &  87.66 \\ 											
    ~ & +EASPP	& $\checkmark$		& 		&  	&  	 & 	88.89 \\ 												
    ~ & +RCCA	& 	 		& $\checkmark$	& 		&   & 89.25 \\ 					 							
    ~ & EffCDNet & $\checkmark$		& $\checkmark$	&   	&  	& 89.94 \\ 				
    ~ & +IEL1	& $\checkmark$		& $\checkmark$	& $\checkmark$	&   & 90.88 \\ 	
    ~ & +IEW	&  $\checkmark$	& $\checkmark$	&  & $\checkmark$ & 91.29 \\ 	
    \hline
  \end{tabular}
\end{table}

\begin{table}[t]
  \scriptsize
  \renewcommand{\arraystretch}{1.3}
  \caption{The F1 score of symmetric architecture and our asymmetric architecture on both datasets}
  \label{table:7}
  \centering
  \begin{tabular}{c c c c c}
    \hline
    \bfseries Architecture	& \bfseries Param.($M$) & \bfseries Layers &  \bfseries	SVCD &  \bfseries	BCD	\\
    \hline\hline
    Backbone-Ours		& 1.48 & 110 & 91.73 &  87.66	 \\ 					 		
    Backbone-Sym		& 1.91 & 111 & 91.20 &  86.80 \\ 					 							
    \hline		
  \end{tabular}
\end{table}

\subsubsection{Ablation Study}\label{sec:3.4.2}

\par In this subsection, to ascertain the contribution of key components in EffCDNet to the overall performance, we conduct a series of ablation experiments on both datasets. The relevant results are reported in Table \ref{table:4}; here, “Backbone” denotes a network with the very deep encoder and shallow decoder, but without EASPP, RCCA, and information entropy loss.

\par It is apparent that, even without the two modules and IE loss, our presented backbone achieves F1 results of 91.73$\%$ and 87.66$\%$ on the SVCD and BCD datasets, respectively, which is comparable to the results achieved by the state-of-the-art method IFN. Then, we also modify our backbone into a symmetric architecture with similar depth and compare it with our asymmetric architecture in Table \ref{table:7}. We can see that the performance of asymmetric architecture is better even though the symmetric architecture has more parameters. This result, to some extent, verifies our claim that symmetric architecture is not necessary, and that network architecture with a very deep encoder and a simple decoder is more suitable for binary change detection tasks. 

\par With the help of EASPP, the F1 is increased by 1.99$\%$ and 1.23$\%$, which indicates the significance of multi-scale change information extraction. Enlarging the feature distance between changed and unchanged pixels by considering non-local similar difference information, RCCA increases the F1 improvement from 91.73$\%$ to 94.30$\%$ on the SVCD dataset and from 87.66$\%$ to 89.25$\%$ on the BCD dataset. When both modules are combines, the network attains an F1 of 94.63$\%$ F1 and 89.94$\%$ F1 on the SVCD and BCD datasets, respectively. 

\par Finally, although the EffCDNet with cross-entropy loss achieves very high F1 (94.63$\%$) on the SVCD dataset, it still achieves performance gains of around 0.6$\%$ after using IEL1 for training. On the BCD dataset, this performance improvement is more obvious: IEL1 and IEW can contribute an F1 increase of 0.94$\%$ and 1.35$\%$, respectively. 

\begin{figure}[!t]
  \centering
  \includegraphics[scale=0.45]{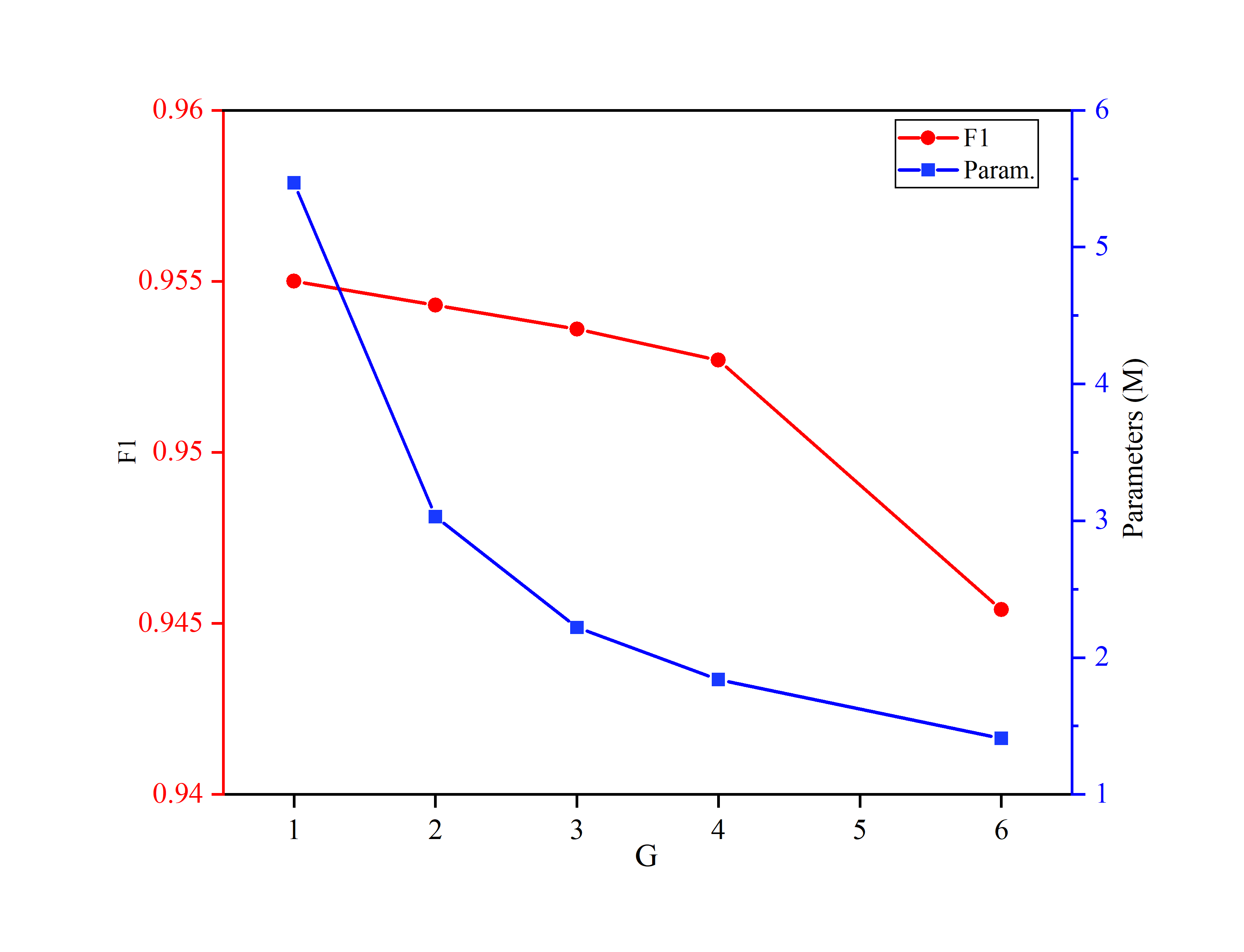}
  \caption{The relationship between the performance of EffCDNet and group number $G$ on the SVCD dataset.}
  \label{fig:15}
\end{figure} 

\begin{figure*}[t]
  \centering
  \subfloat[]{
    \includegraphics[scale=0.44]{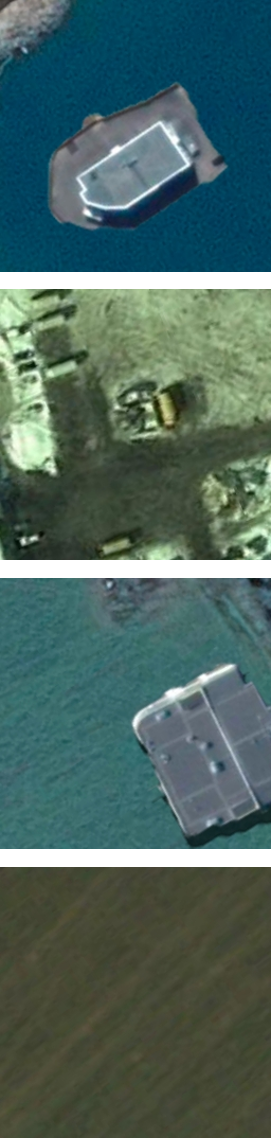}
  \label{fig:CSCD}}
  \subfloat[]{
    \includegraphics[scale=0.44]{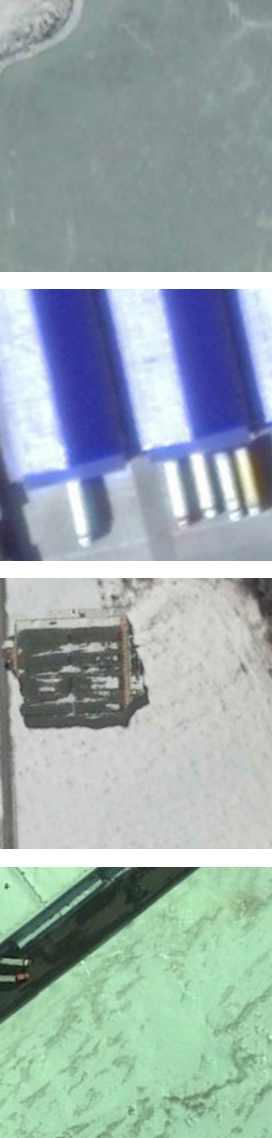}
  \label{fig:CSCD}}
  \subfloat[]{
    \includegraphics[scale=0.44]{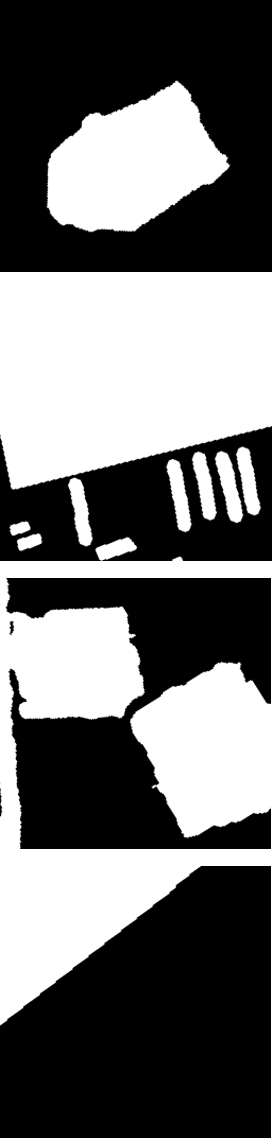}
  \label{fig:moti}}
  \subfloat[]{
    \includegraphics[scale=0.44]{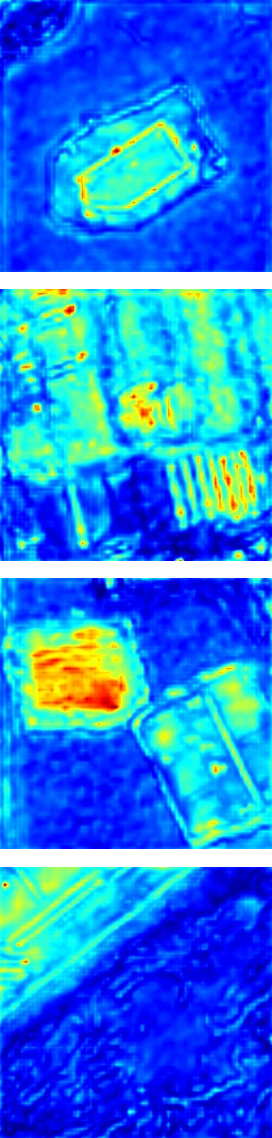}
  \label{fig:moti}}
  \subfloat[]{
    \includegraphics[scale=0.44]{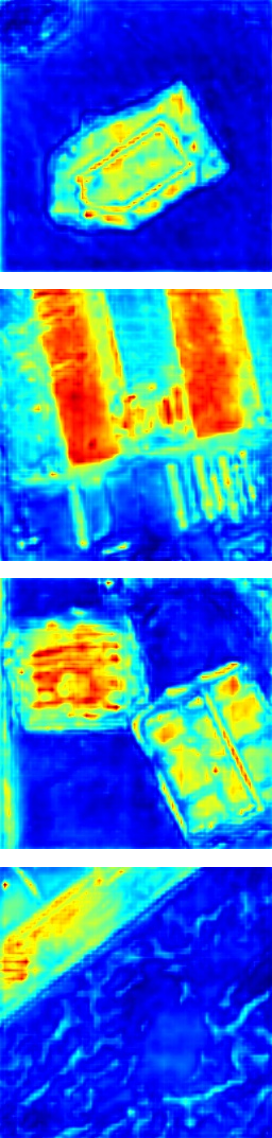}
  \label{fig:moti}}
  \subfloat[]{
    \includegraphics[scale=0.44]{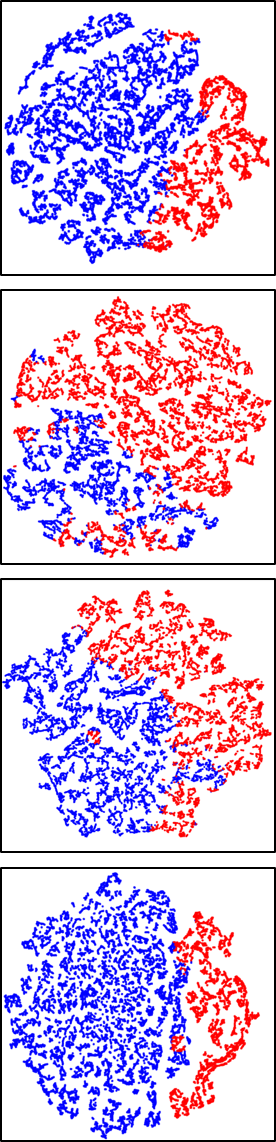}
  \label{fig:moti}}
  \subfloat[]{
    \includegraphics[scale=0.44]{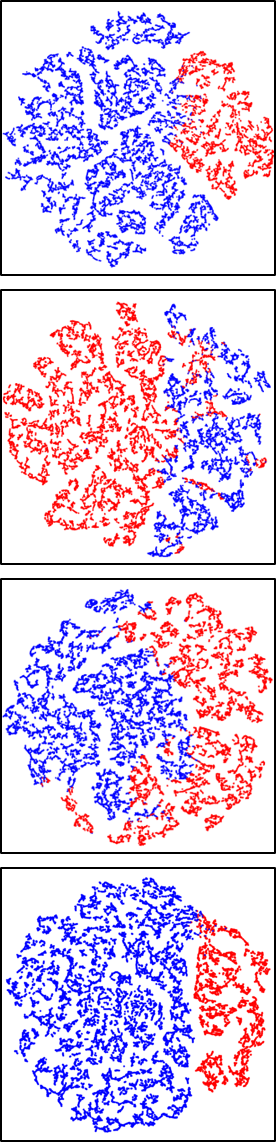}
  \label{fig:moti}}
  \caption{Visualization of deep features before classifier. (a) Pre-change images. (b) Post-change images. (c) Ground truth. (d) Feature distance maps produced by our backbone network. (e) Feature distance maps produced by our backbone network with the RCCA module. (f) t-SNE results of our backbone network. (g) t-SNE results of our backbone network with the RCCA module. In the t-SNE result, red indicates a changed pixel and blue indicates an unchanged pixel.}
  \label{fig:13}
\end{figure*}

\subsubsection{Effects of Group Number}
The group number $G$ of efficient convolution is an important parameter that controls our model size. Therefore, we evaluate the performance of EffCDNet with different values of $G$ on the SVCD dataset, as shown in Fig. \ref{fig:15}. We could see that by utilizing group convolution to reduce model size, the number of parameters reduce largely. Owing to the channel shuffle mechanism, which can ensure the information exchange between different groups, the performance of our approach only see a marginal decline first. However, when $G$ increases to 6, F1 score sees a significant decrease. Therefore, $G=4$ is a good trade-off between the performance of change detection and parameter numbers. 

\subsubsection{Effects of RCCA module}
\begin{table}[t]
    \scriptsize
  
    \renewcommand{\arraystretch}{1.2}
    \caption{Comparison of RCCA module and standard self-attention module on the SVCD dataset. FLOPs are estimated under one image-pair with size of 2$\times$3$\times$256$\times$256}
    \label{table:5}
    \centering
    \begin{tabular}{c c c c}
      \hline
      \bfseries Attention type	& \bfseries R & \bfseries	$\Delta$FLOPs(G) & \bfseries	F1 ($\%$)	\\
      \hline\hline
      -	& -	& 0		& 91.83  \\ 
      \hline											
      \multirow{3}{*}{RCCA}	& 1		& 0.48		& 93.32	 \\ 												
      ~	& 	2		& 0.95 & 94.30	 \\ 					 							
      ~  &  3		& 1.43	 	& 	94.45 \\ 				
      \hline
      Standard	& -	& 10.34	& 94.39\\  										
      \hline
    \end{tabular}
  \end{table}

\par In the decoder of EffCDNet, the RCCA module is used to enlarge the feature distance between changed and unchanged pixels, after which the features pass through the classifier to generate the final result. To qualitatively reflect the effectiveness of the RCCA module, for the feature map before the classifier, we calculate its $L_{2}$ norm to obtain the feature distance map and use the t-SNE algorithm \citep{maaten2008visualizing} to show the distribution of changed and unchanged pixels. As shown in Fig. \ref{fig:13}, compared to the feature distance map of the backbone network (see Fig. 13-(d)), the map yielded by the network using the RCCA module (see Fig. 13-(e)) is more discriminative; here, the discrepancy between changed and unchanged pixels is very obvious, as the changed pixels are clearly highlighted while the unchanged pixels are well suppressed. Moreover, in the t-SNE results, we can see that with the help of the RCCA module, the changed and unchanged pixels are well clustered and have higher internal compactness; in addition, the two types of pixels are significantly separated from each other. Consequently, through the RCCA module, the feature representation at each position achieves improved discriminability from similar feature representations at any position, thereby producing highly discriminative features.

\par Furthermore, we study the relationship between $R$ and model performance and compare the RCCA module with the standard self-attention module. As reported in Table \ref{table:5}, when $R=1$, capturing the relationship between each pixel and the pixels in its criss-cross path can bring about a 1.50$\%$ F1 improvement. By repeating the module twice to capture the dense relationships, the F1 is increased by 0.98$\%$. However, increasing $R$ from 2 to 3 brings about only a slight performance increment. Therefore, as a tradeoff between computational cost and accuracy, $R$ is set to 2 in EffCDNet. More importantly, when $R=2$, the RCCA module shows competitive performance with standard self-attention module, but significantly reduces FLOPs by about 91.8$\%$; this means that, compared to the standard self-attention module, the RCCA module is capable of improving feature discriminability with compared to the standard self-attention module, the RCCA module is capable of improving feature discriminability with more efficient way.

\subsubsection{Effects of Information Entropy Loss}

\begin{figure}[!t]
  \centering
  \subfloat[]{
  \includegraphics[scale=0.58]{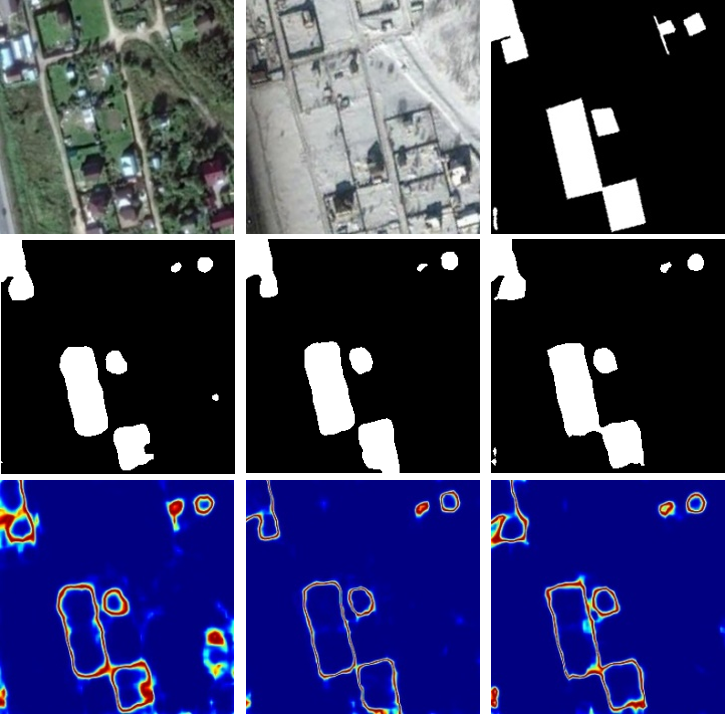}
  }
  \hfil
  \subfloat[]{
  \includegraphics[scale=0.58]{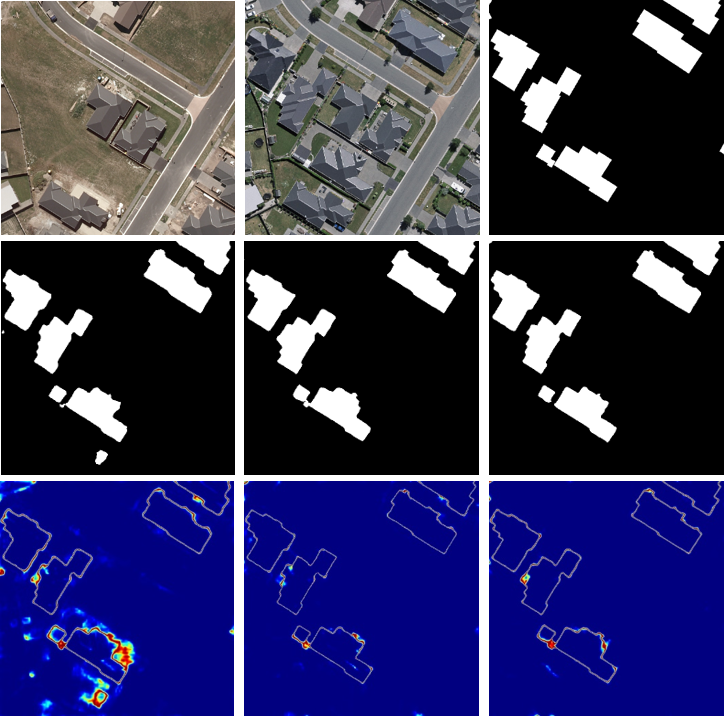}
  }
  \caption{Change maps and entropy maps on the two datasets. (a) Change detection results and corresponding entropy maps on the SVCD dataset. (b) Change detection results and corresponding entropy map on the BCD dataset. From top to bottom: multi-temporal image-pair and ground truth, change maps and corresponding entropy maps obtained by EffCDNet trained with cross-entropy loss, IEL1 and IEW, respectively.}
  \label{fig:14}
\end{figure}

\begin{table}[t]
  \scriptsize

  \renewcommand{\arraystretch}{1.2}
  \caption{The performance of FC-EF and IFN trained by different loss functions on the BCD dataset}
  \label{table:6}
  \centering
  \begin{tabular}{c c c}
    \hline
    \bfseries Model	& \bfseries OA($\%$) &  \bfseries	F1($\%$)	\\
    \hline\hline
    FCEF-CE		& 95.46 & 77.10	 \\ 					 		
    FCEF-IEL1		& 96.46 & 80.41	 \\ 					 							
    FCEF-IEW		& 96.47	 & 	79.84 \\ 		
    \hline		
    IFN-CE		& 97.67 & 87.27	 \\ 					 		
    IFN-IEL1	& 97.80 & 87.83 \\ 					 							
    IFN-IEW		& 97.82 & 87.79 \\ 		
    \hline
  \end{tabular}
\end{table}

\par In section 2.2.4, two novel entropy information loss functions are proposed to solve the problem of cross-entropy loss in optimizing confused pixels. As Table \ref{table:4} shows, the two loss functions can improve network performance. To further intuitively demonstrate the role of information entropy loss, in Fig. \ref{fig:14}, we present the change maps and corresponding entropy maps generated by EffCDNet trained by cross-entropy loss and the two information entropy losses. As this figure shows, the entropy maps of EffCDNet trained by information loss are visually superior, as mere category boundaries have high entropy values while category internal regions have very low entropy values. Moreover, the obtained changed maps match the ground truth very well. However, in the entropy map produced by the network trained by cross-entropy, a proportion of the pseudo-change pixels have high entropy values, indicating that this type of pixel has high uncertainty and may not be well optimized in the training stage. In the corresponding change maps, these high-entropy pixels belonging to the non-change class are misclassified as changed pixels. Consequently, compared to cross-entropy loss, the proposed information entropy-based loss functions are better able to optimize confused pixels during the training stage, thereby improving the network performance in predicting these pixels. 

\par Besides, Table \ref{table:6} presents the performance of the benchmark method FC-EF and the state-of-the-art method IFN trained on the BCD dataset with different loss functions. As is clear, being trained by two information entropy loss functions, FC-EF and IFN achieve better change detection performance. However, the performance improvement of IFN by information entropy loss is not so significant because it has used the deep supervision techinique and dice loss function. The results reported in Table \ref{table:6} indicates that the proposed information entropy losses are not only suitable for the proposed EffCDNet, which is a general loss function and can be used to aid with other change detection network training. In addition, this approach of using information entropy to help in optimizing uncertain samples is not limited to change detection, and can also be employed in other remote sensing image interpretation tasks.

\section{Conclusion}\label{sec:4}
\par Focusing on the major issues with the existing FCN-based change detection network, in this paper, we present the first attempt at designing an end-to-end efficient deep network, called EffCDNet. To obtain excellent change detection performance, EffCDNet is designed to be very deep. In terms of the specific network architecture, EffCDNet adopts an architecture that has a very deep encoder and a lightweight decoder. In order to overcome the problems of numerous parameters and unbearable computational cost brought about by deep architecture, in EffCDNet, almost all standard convolution layers are replaced by an efficient convolution. Moreover, the design and application of the key components of EffCDNet also give consideration to both performance boost and computational overhead. Consequently, although the proposed EffCDNet achieves a depth of 116 layers, which is far deeper than existing methods, its number of parameters is only slightly higher than benchmark models, while its computational cost is much lower than the state-of-the-art models. In addition, we present two information entropy-based loss functions to optimize EffCDNet. Compared with cross-entropy loss, the proposed information entropy-based loss function can better optimize confused pixels, thereby improving the network performance. Detailed experiments conducted on two challenging open change detection datasets demonstrate the effectiveness and superiority of our approach.

\section*{References}

\bibliography{EffCDNet}

\end{document}